\documentclass[lettersize,journal]{IEEEtran}
\usepackage{amsmath,amsfonts,amssymb}
\usepackage{algorithmic}
\usepackage{array}
\usepackage[caption=false,font=normalsize,labelfont=sf,textfont=sf]{subfig}
\usepackage{textcomp}
\usepackage{stfloats}
\usepackage{float}
\usepackage{url}
\usepackage{verbatim}
\usepackage{graphicx}
\usepackage{epstopdf}
\usepackage{adjustbox}
\usepackage{booktabs}
\usepackage{multirow}
\usepackage{subcaption}
\usepackage{utfsym}

\def\BibTeX{{\rm B\kern-.05em{\sc i\kern-.025em b}\kern-.08em
    T\kern-.1667em\lower.7ex\hbox{E}\kern-.125emX}}
\usepackage{balance}

\begin{document}

\title{VFM-ISRefiner: Towards Better Adapting Vision Foundation Models for Interactive Segmentation of Remote Sensing Images}
\author{Deliang Wang, Peng Liu, Yan Ma, Rongkai Zhuang, Lajiao Chen, Bing Li, and Yi Zeng
\thanks{This research was funded by National Key Research and Development Program of China under Grant 2024YFF1307204, Comprehensive Site Selection System Project grant number E5E2180501, and National Natural Science Foundation of China under Grant U2243222.

Deliang Wang, Rongkai Zhuang and Bing Li are with Aerospace Information Research Institute, Chinese Academy of Sciences, Beijing 100094, China, and the School of Electronic, Electrical and Communication Engineering, University of Chinese Academy of Sciences, Beijing 101408, China.

Peng Liu, Yan Ma and Lajiao Chen are with Aerospace Information Research Institute, Chinese Academy of Sciences, Beijing 100094, China (e-mail: liupeng202303@aircas.ac.cn).

Yi Zeng is with the School of Information Science and Technology, Beijing Forestry University, Beijing 100083, China.

}}

\markboth{Journal of \LaTeX\ Class Files,~Vol.~18, No.~9, September~2020}%
{How to Use the IEEEtran \LaTeX \ Templates}

\maketitle

\begin{abstract}
Interactive image segmentation(IIS) plays a critical role in generating precise annotations for remote sensing imagery, where objects often exhibit scale variations, irregular boundaries and complex backgrounds. However, existing IIS methods, primarily designed for natural images, struggle to generalize to remote sensing domains due to limited annotated data and computational overhead. To address these challenges, we propose VFM-ISRefiner, a novel click-based IIS framework tailored for remote sensing images. The framework employs an adapter-based tuning strategy that preserves the general representations of Vision Foundation Models while enabling efficient learning of remote-sensing–specific spatial and boundary characteristics. A hybrid attention mechanism integrating convolutional local modeling with Transformer-based global reasoning enhances robustness against scale diversity and scene complexity. Furthermore, an improved probability map modulation scheme effectively incorporates historical user interactions, yielding more stable iterative refinement and higher boundary accuracy. Comprehensive experiments on six remote sensing datasets, including iSAID, ISPRS Potsdam, SandBar, NWPU, LoveDA Urban and WHUBuilding, demonstrate that VFM-ISRefiner consistently outperforms state-of-the-art IIS methods in terms of segmentation accuracy, efficiency and interaction cost. These results confirm the effectiveness and generalizability of our framework, making it highly suitable for high-quality instance segmentation in practical remote sensing scenarios. The codes are available at https://github.com/wondelyan/VFM-ISRefiner .
\end{abstract}



\section{Introduction}
\IEEEPARstart{I}{nteractive} Image Segmentation(IIS) is an effective approach that enables high-precision, pixel-level annotation with minimal human interaction. By continuously incorporating user feedback during the segmentations such as bounding boxes\cite{xu2017deep}, clicks\cite{sofiiuk2022reviving}\cite{jang2019interactive} and scribbles\cite{wu2014milcut}, IIS can accurately delineate object boundaries and has been widely applied in tasks including medical image analysis, natural scene understanding and remote sensing image interpretation. Compared with fully automatic segmentation methods, IIS offers higher controllability and flexibility, demonstrating significant advantages when handling ambiguous or fine-grained targets\cite{liu2022survey}. For example, in remote sensing imagery, IIS can effectively distinguish shadow-covered buildings from adjacent vegetation or identify irregularly shaped sandbars within rivers, with only a small amount of user interaction required to correct model prediction errors.

\begin{figure}[t]
    \centering
    \includegraphics[scale=0.3]{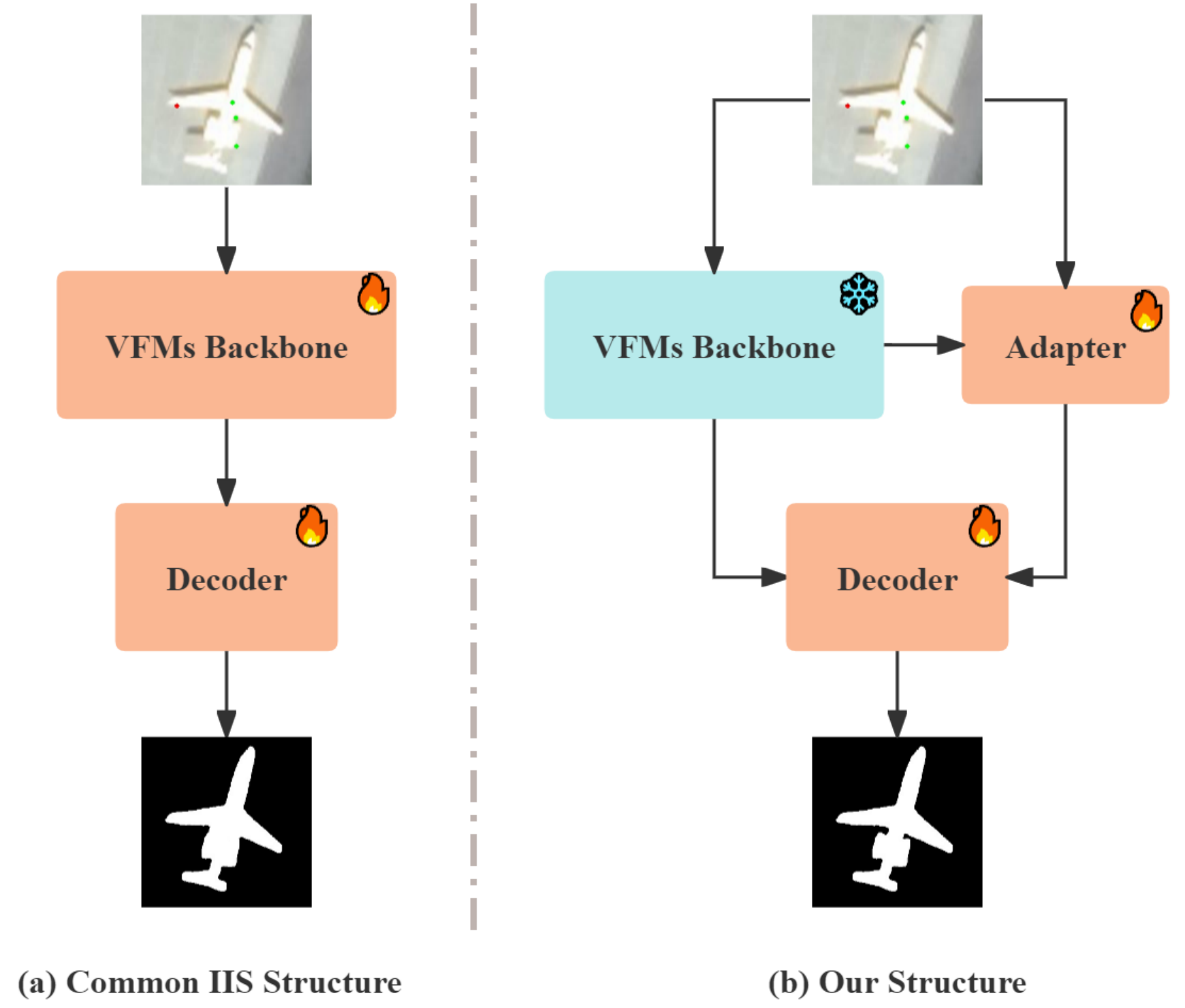}
    \caption{\textbf{Structure Comparison.} (a) represents the common IIS structure. (b) is the simplified structure used in our paper.}
    \label{Introduction_Image}
\end{figure}

In recent years, the emergence of Vision Foundation Models (VFMs) has ushered in new opportunities for the field of Interactive Image Segmentation (IIS). Vision Foundation Models (VFMs) are typically built upon large-scale pretrained Convolutional Neural Networks (CNNs) and Transformers (e.g., ResNet\cite{he2016deep}, ConvNeXt\cite{liu2022convnet}, ViT\cite{dosovitskiy2020image}, and Swin Transformer\cite{liu2021swin}), possessing rich transferable representations learned from massive and diverse datasets\cite{song2025refining}. Currently, in the field of IIS, numerous studies such as BRS\cite{jang2019interactive}, RITM\cite{sofiiuk2022reviving} and SimpleClick\cite{liu2023simpleclick} adopt VFMs as feature extraction encoders, and design additional downstream decoders to complete feature parsing and segmentation mask generation, as showed in Fig.\ref{Introduction_Image}(a). However, most existing VFMs serving as feature encoders are pretrained exclusively on large-scale natural images, and their performance degrades significantly when directly applied to remote sensing images without further training. For instance, although SAM\cite{kirillov2023segment} and its variants demonstrate superior performance on natural images, they require numerous corrective clicks to handle RS scenarios such as small vehicles segmentation(sensitive to scale variation) or shadow-covered buildings(with noisy boundaries), leading to substantially increased interaction costs\cite{OSCO2023103540}. To address this issue, most existing IIS methods perform task-specific full-parameter fine-tuning on the encoders of VFMs for interactive segmentation of remote sensing targets. Nevertheless, training VFMs consumes extensive computational resources, and full-parameter fine-tuning may overwrite the general visual knowledge acquired during the pre-training phase, thereby reducing the models' inherent generalization capability\cite{han2023survey}. Additionally, annotated data in the remote sensing domain is scarce and costly to obtain\cite{zhang2023new}, which limits the improvement of model's generalization in this specific domain\cite{wang2023interactive}.

With the advancement of multimodal large models, several fine-tuning adaptation algorithms (e.g., Adapter\cite{yin2024parameter} and Prompt\cite{lester2021power}) have also evolved. By leveraging fine-tuning mechanisms, pretrained backbone networks can be efficiently transferred to downstream tasks, reducing training costs while preserving general visual knowledge\cite{awais2025foundation}. For interactive segmentation in remote sensing, leveraging VFMs enables the preservation of general visual priors while selectively optimizing high-level features to better align with the spatial characteristic of RS imagery, as showed in Fig.\ref{Introduction_Image}(b). For instance, adapter modules can be employed to capture domain-specific edge structures of RS targets such as the linear boundaries of building rooftops or the elongated contours of roads, without altering the parameters of the pretrained backbone network\cite{rs17030369}.

However, existing IIS frameworks still face two major limitations when applied to remote sensing imagery. First, most methods(e.g., SimpleClick\cite{liu2023simpleclick} and AdaptiveClick\cite{lin2024adaptiveclick}) fail to fully exploit the historical information of user interactions. Although some approaches(e.g., MFP\cite{lee2024mfp}) attempt to propagate such information through probability map modulation, conventional modulation mechanisms with fixed circular modulation windows are not well-suited to the irregular shapes of remote sensing targets. As a result, prediction instability or oscillation often occurs during multi-round interactions, particularly when dealing with geometrically complex objects such as river sandbars. Second, although VFMs provide powerful visual priors, directly transferring them to remote sensing imagery often leads to degraded boundary accuracy due to domain discrepancies. The visual features learned from natural images (e.g., object textures and edge details) do not align well with the spectral and spatial characteristics of remote sensing objects (e.g., spectral reflectance and spatial configuration)\cite{chen2024rsprompter}. Moreover, existing adaptation modules such as the Adapter in SAM-HQ\cite{ke2023segment} are not specifically designed for remote sensing scenarios, making it difficult to capture the intrinsic spectral-spatial correlations of ground objects\cite{han2023survey}. Therefore, there is an urgent need for an IIS framework that integrates an efficient VFM-based adaptation mechanism tailored for remote sensing with a robust historical information propagation strategy, enabling high-precision segmentation with minimal user interaction cost.

In response to the above issues, we proposed a novel click-based IIS framework for remote sensing images: VFM-ISRefiner, incorporating the probability map modulation mechanism and the backbone-freezing adapter fine-tuning algorithm to address the unique challenges of remote sensing data. Drawing on the insights of MFP\cite{lee2024mfp}, we leverage probability map modulation to propagate historical interaction information, thereby enhancing the model's capability to retain object shapes and spatial contexts during iterative processes. With respect to the design of the adapter fine-tuning algorithm, we introduce a module to capture local spatial correlations and edge features specific to remote sensing objects, eliminating the need to retrain the entire backbone network. 

In summary, our work advances remote sensing IIS by:

1. Designing and integrating an adapter module to enhance Vision Foundation Models, which preserves the model's pretrained robust feature representations while enabling the learning of remote sensing-specific local details. This module leverages the generalized feature extraction capabilities of foundation models to reduce training costs, while acquiring domain-adaptive feature perception abilities to improve the precision of object edges in remote sensing scenarios.

2. Designing a composite attention mechanism that integrates convolutional networks with transformer-based attention to handle scale variations and complex backgrounds in remote sensing images. This hybrid framework leverages convolutional networks for local spatial modeling and attention mechanisms for global contextual reasoning, enhancing adaptability to diverse scales and cluttered backgrounds in remote sensing scenarios.

3. Adapting an enhanced probability map modulation algorithm to propagate historical interaction information, thereby reinforcing the retention of object shape and spatial context in remote sensing scenarios.


\section{Related Work}
\subsection{Interactive Image Segmentation}
Interactive image segmentation (IIS) has evolved significantly over the past decades, with advancements spanning traditional optimization, deep learning and transformer-based framework. Early interactive segmentation methods relied on handcrafted features and graph-based optimization. Techniques such as GrabCut\cite{rother2004grabcut} used bounding boxes or scribbles to initialize foreground-background separation via minimizing energy function, while Random Walk\cite{grady2006random} and Graph Cuts\cite{boykov2001interactive} modeled segmentation as a graph partitioning problem, leveraging pixel similarity metrics. These methods, however, struggled with complex scenes due to their reliance on shallow features, limiting their applicability to high-resolution or noisy remote sensing images.

The rise of deep learning revolutionized IIS. DIOS\cite{xu2016deep} first introduced convolutional neural networks (CNNs) for interactive object selection, encoding user clicks into distance-transformed "click maps" and fusing them with RGB images as network input. This paradigm became the foundation of follow-ups, with subsequent works refining network architectures and interaction strategies. For example, BRS\cite{jang2019interactive} and f-BRS\cite{sofiiuk2020f} introduced backpropagation-based refinement to correct mislabel clicks during inference, improving accuracy but increasing computational overhead. As deep learning matured, researchers exploring the integration of prior segmentation results to stabilize predictions. RITM\cite{sofiiuk2022reviving} demonstrated that feeding previous masks into the network enhanced model stability, as iterative clicks often target mislabeled regions from prior rounds. This insight led to widespread adoption of "history information" as input, forming a pipeline for click-based IIS\cite{chen2022focalclick}\cite{lin2020interactive}\cite{du2023efficient}\cite{lin2022focuscut}.

The shift to transformer architectures further elevated IIS performance. ViT\cite{dosovitskiy2020image} and its variants, with their strong feature modeling capabilities, surpassed CNNs in capturing long-range spatial dependencies which are critical for segmenting large or irregular objects. SimpleClick\cite{liu2023simpleclick} was the first to adopt a plain ViT backbone for IIS, outperforming CNN-based methods by leveraging pretrained transformer weights to encode rich contextual information. Based on this, follow-ups improved the backbone, the encoder and other structures\cite{sun2024cfr}\cite{yan2023piclick}\cite{lin2024adaptiveclick}\cite{xu2025mst}\cite{zhang2025ntclick}. Despite these advances, most approaches still rely on retraining large models, posing challenges for computational efficiency.

Although Transformer-based IIS frameworks have significantly improved segmentation performance, their adaptation to remote sensing imagery still requires costly full fine-tuning and large annotated datasets. This limitation motivates the exploration of Vision Foundation Models (VFMs), which leverage large-scale pretrained Transformers such as ViT and can be efficiently adapted through fine-tuning strategies for interactive segmentation tasks.

\subsection{Vision Foundation Models and Parameter-efficient Fine-Tuning}
Vision Foundation Models have become the driving force of modern computer vision. Early convolutional neural networks (CNNs), such as FCN\cite{long2015fully}, VGG\cite{simonyan2014very}, ResNet\cite{he2016deep}, DenseNet\cite{huang2017densely} and HRNet\cite{sun2019deep}, provided effective feature extraction for pixel-level tasks, while the introduction of Transformers\cite{vaswani2017attention}, such as ViT\cite{dosovitskiy2020image} and Swin Transformer\cite{liu2021swin}, significantly enhanced global modeling and multi-scale representation. More recently, with the advent of large-scale pretraining, models including MAE\cite{he2022masked}, SAM\cite{kirillov2023segment}, DINOv2\cite{oquab2023dinov2} and ConvNeXt\cite{liu2022convnet} have demonstrated remarkable generalization across tasks and domains. These models have achieved state-of-the-art performance in classification, detection and segmentation, establishing themselves as universal feature extractors with strong potential for remote sensing and interactive segmentation\cite{liang2025vision}\cite{lu2025vision}. 

Nevertheless, despite their representational power, directly applying VFMs to interactive segmentation in remote sensing remains challenging due to high computational costs. To bridge this gap, a range of parameter-efficient fine-tuning (PEFT) strategies have been developed\cite{zhang2025high}. Representative PEFT approaches include Adapter Tuning\cite{yin2024parameter}, Prompt Tuning\cite{lester2021power} and LoRA\cite{hu2022lora}, which reduce training overhead by introducing trainable modules while keeping the pretrained backbone largely frozen. Prompt Tuning requires minimal parameters but often offers limited flexibility for complex visual tasks. LoRA achieves high efficiency and incurs no inference overhead\cite{hu2022lora}, but its low-rank formulation can restrict representational capacity in domains with large distribution gaps such as natural images to RS images. In contrast, Adapter Tuning strikes a favorable balance between efficiency and expressiveness, enabling tasks-specific adaptation with relatively small parameter costs while maintaining robust performance\cite{yin2024parameter}.

Several studies have demonstrated the promise of adapter-based designs in IIS. AdaptFormer\cite{chen2022adaptformer} improves visual adaptability by integrating adapters into transformer blocks, while SAM-HQ\cite{ke2023segment} and SAM2Refiner\cite{yao2025towards} employ adapter-driven refinement to enhance mask quality without retraining the entire backbone. ClickAdapter\cite{li2024clickadapter} further incorporates user-interaction cues directly into adapter modules. Nevertheless, these approaches remain primarily tailored for natural imagery and do not explicitly model domain-specific properties such as spectral heterogeneity, irregular boundaries, or spatial complexity inherent to remote sensing data. This motivates the need for adapter mechanisms that preserve VFM generalization while enabling domain-sensitive adaptation for remote sensing IIS.

\subsection{Historical Information Utilization in Interactive Refinement}
Apart from architectural adaptation, IIS performance heavily depends on how effectively a model leverages user's historical interactions during iterative refinement. RITM\cite{sofiiuk2022reviving} and f-BRS\cite{sofiiuk2020f} pioneered the use of previous segmentation masks as network input in the current segmentation round. The two models showed that even a simple feed-forward model using previous masks can enhance model stability without additional optimization. Since then, interactive segmentation methods have all adopted previous segmentation masks as network input. FocalClick\cite{chen2022focalclick} and FocusCut\cite{lin2022focuscut} refined results within a local window by correcting global predictions with local ones around clicks. GPCIS\cite{zhou2023interactive} formulated the problem of interactive segmentation as a Gaussian process classification task. EMC-Click\cite{du2023efficient} used self-attention and correlation modules to spread click information to unclicked areas. MST\cite{xu2025mst} addressed the issue of target scale variations in interactive segmentation by selectively fusing important multi-scale tokens based on user clicks to enhance the distinguishability between target and background tokens. Although the above methods all use previous masks or previous probability maps in the current round, none of them attempt to extract more effective information from the probability maps. MFP\cite{lee2024mfp} uses gamma correction-based probability map modulation to amplify probabilities near positive clicks and suppress those near negative clicks within a local window, and introduces late fusion of probability-related and backbone features to let historical context influence final prediction. However, the modulation mechanism of MFP\cite{lee2024mfp} struggles with irregularly bounded remote sensing targets, and needs to be improved.


\begin{figure*}[!t]
    \centering
    \includegraphics[scale=0.3]{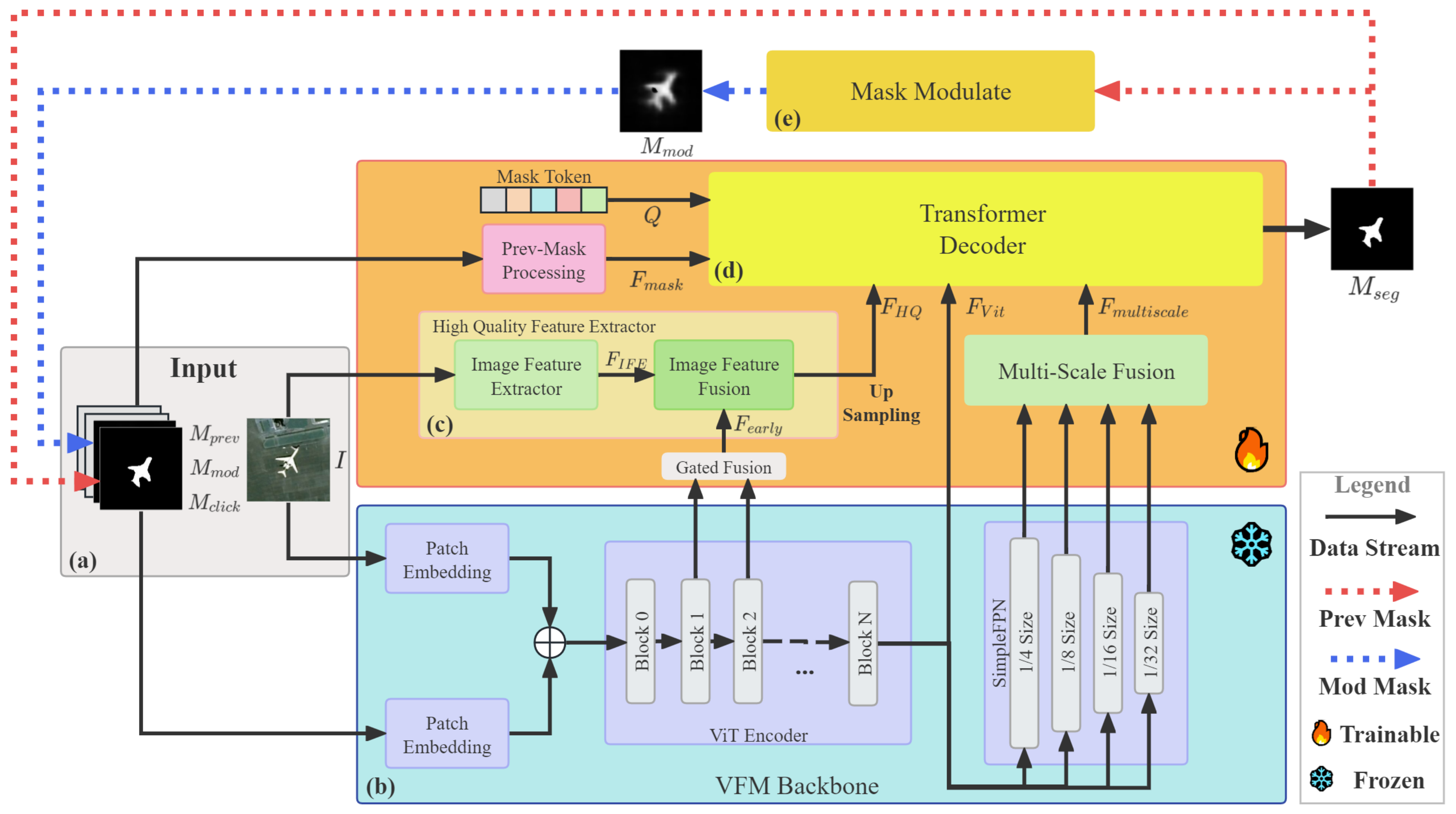}
    \caption{\textbf{Overall Structure.} Our architecture consists of three main components: a frozen Vision Foundation Model (VFM) backbone (blue region), a parameter-trained extractor and decoder (orange-red region) and a probability map modulation module at the top of the diagram. Both the training and testing processes follow a recursive workflow.}
    \label{overall_structure}
\end{figure*}

\section{Method}
In this section, Subsection A will introduce the overall structure of the model as well as the process of feature extraction and transformation; Subsection B will present the basic architecture of the VFM backbone utilized in this study; Subsection C will elaborate on the detailed design of the high-quality feature extractor, explaining how to extract high-quality spatial features from remote sensing images; Subsection D will present the transformer decoder component of the model; and Subsection E will discuss the improvements made to the probability map modulation algorithm.

\subsection{Overall Structure}
To effectively adapt large-scale Vision Foundation Models (VFMs) to remote sensing interactive segmentation tasks, our method adopts a modular backbone-adapter architecture that enables efficient fine-tuning while preserving the pretrained model's generalization capability. As illustrated in Fig.\ref{overall_structure}, the interactive segmentation model proposed in this paper mainly consists of four components: a frozen VFM Backbone, a High-Quality Feature Extractor, a Transformer Decoder and a Probability Map Modulation Module. The framework follows a recursive segmentation-refinement workflow, in which the model iteratively updates its prediction according to user interactions, thereby achieving progressively refined segmentation results with minimal clicks.

\begin{figure}[t]
    \centering
    \includegraphics[scale=0.51]{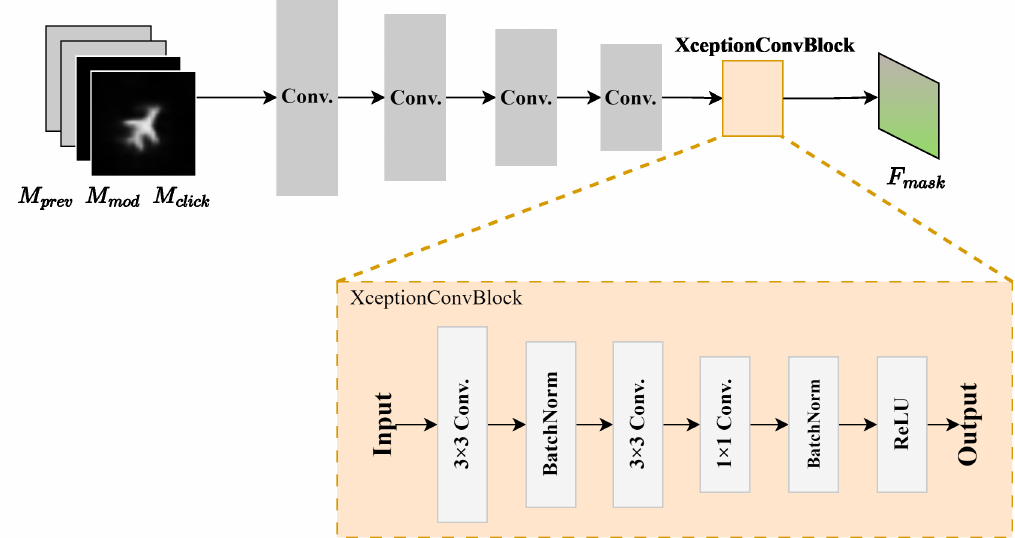}
    \caption{\textbf{Previous Mask Processing Module.} The module is mainly composed of four convolutional modules and one Xception-convolutional module, where the light orange region at the bottom shows the detailed structure of the Xception-convolutional module.}
    \label{Prev_Mask Processing}
\end{figure}

Since this paper adopts the idea of probability map modulation from MFP\cite{lee2024mfp}, in each interaction round, as shown in Fig.\ref{overall_structure}(a), the input to the network here consists of the image $I\in \mathbb{R}^{3\times H\times W}$, the previous segmentation mask $M_{prev.}\in \mathbb{R}^{1\times H\times W}$, the previous modulated segmentation mask $M_{mod.}\in \mathbb{R}^{1\times H\times W}$ and the click maps $M_{click}\in \mathbb{R}^{2\times H\times W}$ generated from the positive and negative clicks set. In the case of the initial click, since $M_{prev.}$ and $M_{mod.}$ do not exist, it is necessary to set both masks as all-zero matrices. Furthermore, as illustrated in Fig.\ref{overall_structure}(b), $M_{prev.}$, $M_{mod.}$ and $M_{click}$ are concatenated into a unified tensor $M\in \mathbb{R}^{4\times H\times W}$ and jointly processed with the image $I$ through the frozen VFM Backbone to obtain stable and semantically rich representations. The High-Quality Feature Extractor presented in Fig.\ref{overall_structure}(c) then complements these backbone features with fine-grained spatial and boundary information, which are fused through cross-attention operations to generate enhanced local-global feature maps. Consistent with the methodology employed in MFP\cite{lee2024mfp}, $M_{prev.}$, $M_{mod.}$ and $I$ are subjected to Conv. and XceptionConv. operations to extract the corresponding probability map feature, which are subsequently fed into the decoder. This paper retains this approach, with the sole modification being that in the input components, image $I$ is substituted with the click map $M_{click}$, as illustrated in Fig.\ref{Prev_Mask Processing}. Research findings from \cite{yan2023piclick} have demonstrated that encoding the click map and transmitting it to the decoder contributes to enhancing the model's localization performance. The three inputs are processed by the Prev-Mask Processing Module, ultimately outputting the mask feature $F_{mask}$. Subsequently, the Transformer Decoder in Fig.\ref{overall_structure}(d) integrates image features, mask features and a learnable mask token to produce an updated instance mask $M_{seg.}$ that balances contextual consistency with edge precision. Finally, as shown in Fig.\ref{overall_structure}(e) the predicted mask undergoes probability map modulation to correct local inconsistencies based on the spatial distribution of user clicks, yielding an updated probability map $M_{mod.}$ that is propagated to the next iteration.

\subsection{Frozen VFM Backbone}

To ensure generalization and efficiency, we employ a frozen Vision Foundation Model pretrained on large-scale natural image datasets as the backbone of our framework. Specifically, we utilize the ViT-Based backbone from MFP\cite{lee2024mfp}, which provides strong semantic and contextual representations. The backbone parameters remain frozen throughout training, preserving its pretrained knowledge while avoiding overfitting on the limited annotated remote sensing data.

As illustrated in Fig.\ref{overall_structure}, the VFM backbone consists of three key parts: Patch Embedding, ViT Encoder and SimpleFPN. The image $I$ and concatenated auxiliary tensor $M$ are first processed by the Patch Embedding layer to obtain embedded vectors, which are then passed through the ViT Encoder blocks to extract hierarchical global features. To incorporate local spatial details, we extract the outputs from early ViT blocks $F_{ViT}^1$ and $F_{ViT}^2$, and fuse them using a learnable gating mechanism:

\begin{equation}
    \label{deqn_ex1}
    \left\{
    \begin{aligned}
        w_{gate} &= \sigma(W\cdot concat\left(F_{ViT}^1,F_{ViT}^2\right)+b) \\
        F_{early} &= w_{gate}\cdot F_{ViT}^1+(1-w_{gate})\cdot F_{ViT}^2
    \end{aligned}
    \right.
\end{equation}

\noindent where $W$ is the weight matrix in the linear transformation. $b$ is the bias term in the linear transformation. $\sigma$ denotes the Sigmoid activation function, which serves to map the result of the linear transformation to the interval $\left[0,1\right]$, thereby obtaining the gating weights. The fused feature $F_{early}$ captures mid-level contextual and spatial details, complementing the high-level semantic features extracted by deeper layers. To address the scale variation inherent in remote sensing imagery, the final ViT block output $F_{ViT}^N$ is fed into the SimpleFPN to generate multi-scale representations ${F_1^{ms}, F_2^{ms}, F_3^{ms}, F_4^{ms}}$, which are fused into a unified multi-scale feature map $F_{multiscale}$. The multi-scale feature will be fed into the trainable transformer decoder, enabling the network to maintain robustness across diverse object sizes.

\subsection{High-Quality Feature Extractor}

While the frozen VFM backbone provides rich global semantics and structural priors, these features alone are insufficient to capture fine-grained local variations in remote sensing targets. Therefore, we design a High-Quality Feature Extractor to complement the VFM outputs with detailed spatial representations. This module consists of two components, namely the Image Feature Extractor and the Image Feature Fusion.

\begin{figure}[h]
    \includegraphics[scale=0.23]{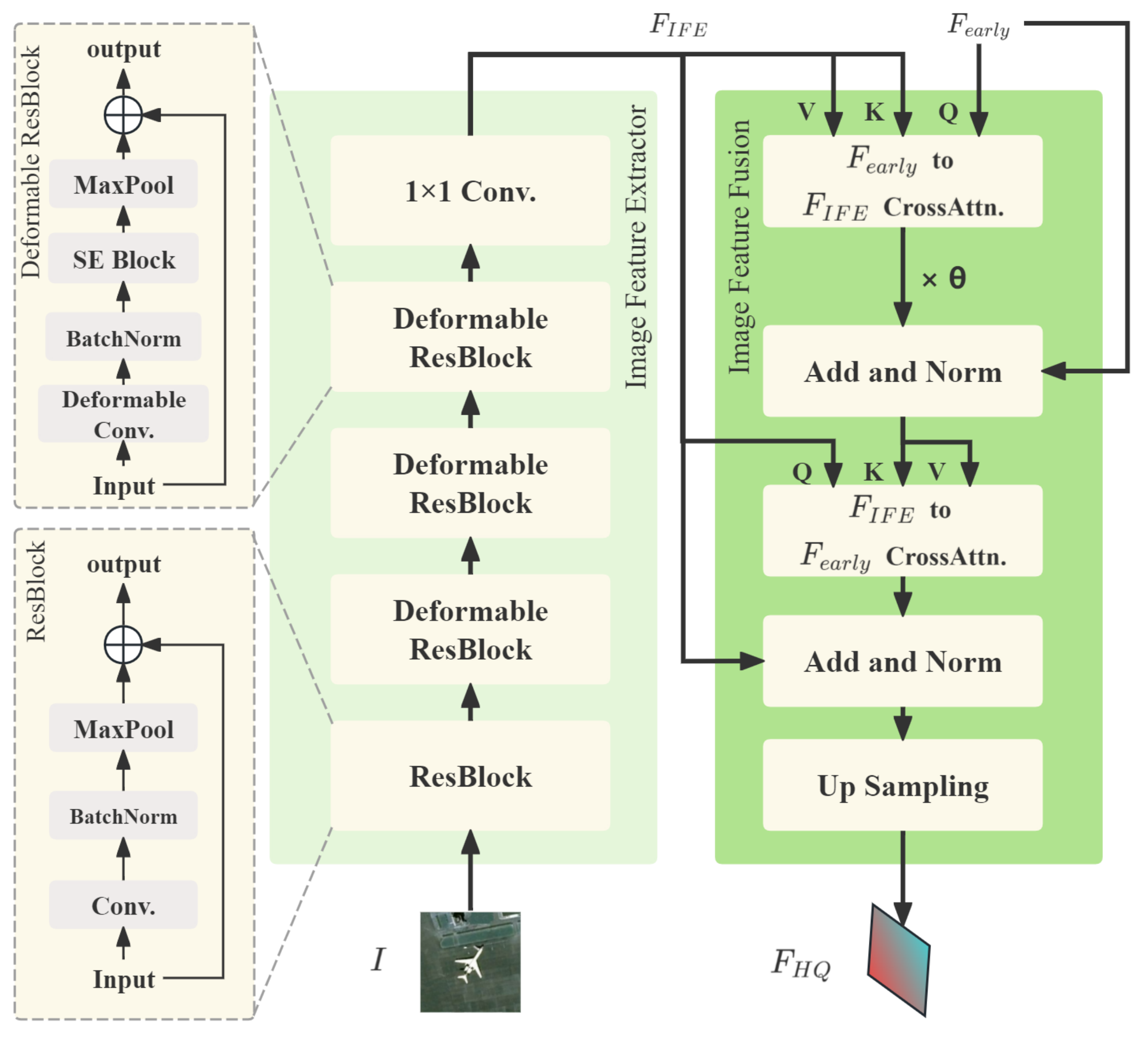}
    \caption{\textbf{High Quality Feature Extractor.} This module comprises an image feature extractor (light-colored region in the middle) and a feature fusion module (green region on the right). The image feature extractor includes one ResBlock, three DeformableResBlocks and one 1×1 convolution, with the detailed architectures of ResBlock and Deformable ResBlock shown on the left of the diagram. The feature fusion module is primarily composed of cross-attention blocks, where $\theta$ serves as a learnable parameter.}
    \label{High_Quality_Feature_Extractor}
\end{figure}

As shown in Fig.\ref{High_Quality_Feature_Extractor}, the Image Feature Extractor adopts a residual network architecture, comprising one convolutional residual module, three deformable convolutional residual modules and one linear projection layer. Herein, deformable convolutions are introduced to learn sampling position offsets, enabling adaptive adaptation to variations in the shape, orientation and scale of remote sensing targets. Moreover, the integration of the Squeeze-and-Excitation (SE) Attention mechanism into the deformable convolutional residual modules serves as a widely adopted strategy in convolutional networks to enhance the feature extraction capability of the network. Upon processing image $I$ through the Image Feature Extractor, a spatially informative feature $F_{IFE}$ is derived.

The feature $F_{IFE}$, together with $F_{early}$ from the early blocks of ViT, is then fed into the Image Feature Fusion module to undergo cross-attention-based fusion. Initially, $F_{early}$ is designated as Query, with $F_{IFE}$ serving as both Key and Value, to execute a $F_{early}$ to $F_{IFE}$ cross-attention operation. The mechanism enables ViT-derived features to integrate partial characteristic information from the convolutional network. Subsequently, the cross-attention output undergoes an element-wise multiplication with a learnable parameter $\theta$ which is initialized separately. The resulting product is then added residually to the original $F_{early}$, followed by a layer normalization operation, as formulated below.
\begin{equation}
    \label{deqn_ex2}
    F_{early}^\prime=LN(F_{early}+\theta\cdot(CA(F_{early},F_{IFE})))
\end{equation}

\noindent where $CA(\cdot)$ is the cross-attention operation, and $LN(\cdot)$ stands for the layer normalization. The reason for introducing the learnable parameter is that the features extracted by ViT contain rich information, and thus this parameter acts to prevent such features from being excessive interference. Subsequently, $F_{IFE}$ is used as Query, while the updated $F_{early}^\prime$ serves as both Key and Value to perform the $F_{IFE}$ to $F_{early}^\prime$ cross-attention operation. This process enables the feature $F_{IFE}$ extracted from convolutional network to assimilate the rich feature information from ViT. The cross-attention output is then added residually to the original $F_{IFE}$, followed by layer normalization. This sequence of operations preserves information integrity and stabilizes the fusion results, as formalized in the equations below.
\begin{equation}
    \label{deqn_ex3}
    F_{IFE}^\prime=LN(F_{IFE}+CA(F_{IFE},F_{early}^\prime))
\end{equation}

Finally, the fused feature $F_{IFE}^\prime$ undergoes shape adjustment and is upsampled to a high resolution, yielding the high-quality image feature $F_{HQ}$.

\subsection{Transformer Decoder}
Built upon the high-level semantic features extracted by the frozen Vision Foundation Model, the decoder establishes effective information interaction between the VFM's global representations and the locally enhanced features from the extractor module. Through bidirectional cross-attention processing, the model establishes efficient information interaction between global semantics and local details, thereby generating prediction results with both global consistency and high-precision boundaries. 

\begin{figure}[h]
    \includegraphics[scale=0.24]{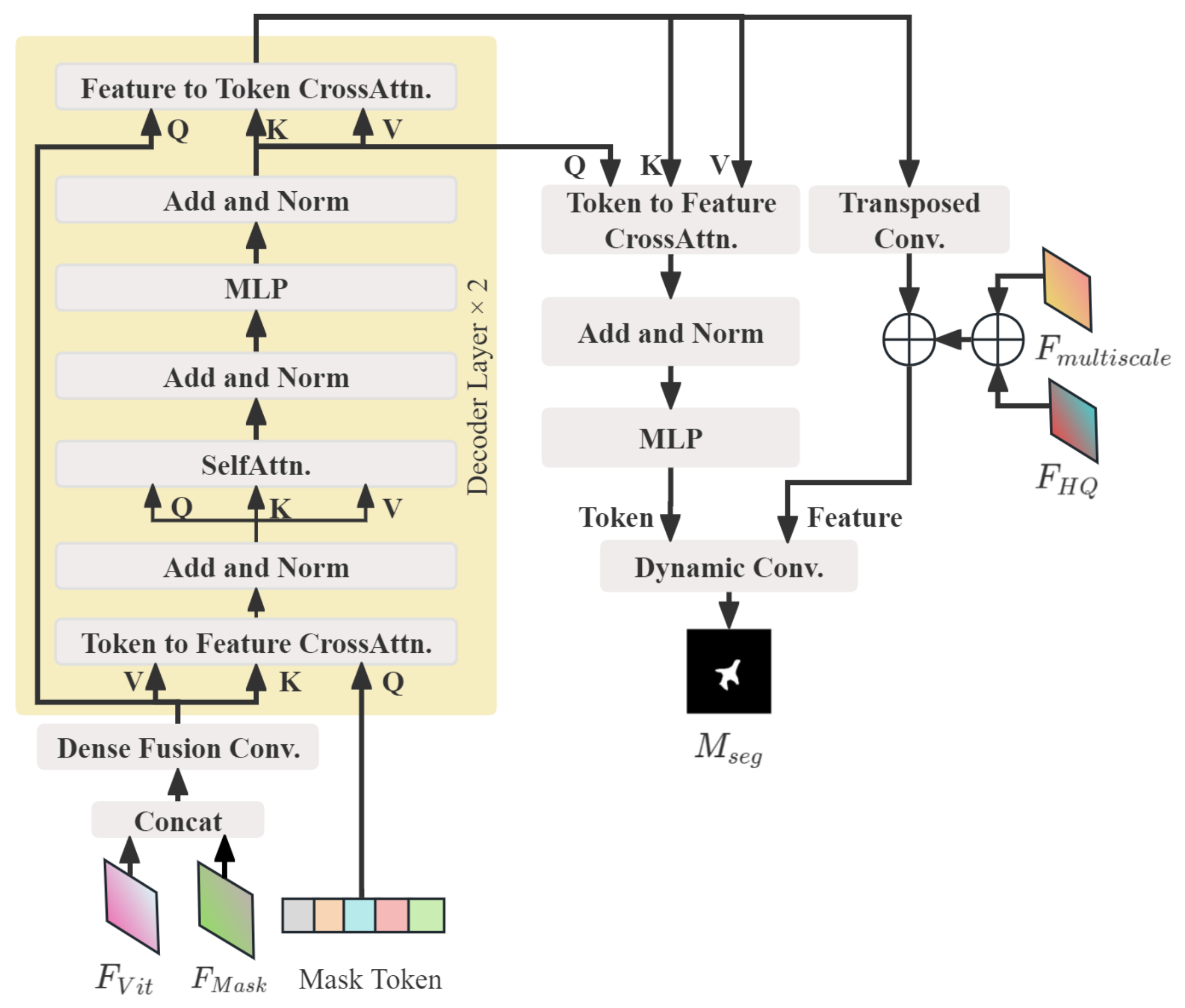}
    \caption{\textbf{Transformer Decoder.} The module consists of a Decoder Layer (the yellow region on the left) and a dual-branch structure (the right region), where the latter includes a Token path and a Feature path. Their outputs are fused via a dynamic convolution module to generate the predicted mask.}
    \label{Transformer_Decoder}
\end{figure}

As illustrated in Fig.\ref{Transformer_Decoder}, First, $F_{mask}$ and $F_{ViT}^N$ are concatenated, then undergo channel-wise fusion via the Dense Fusion Convolution, which primarily consists of $1\times1$ convolution operations, to align dimensions with those of token $Q$. The resulting features $F_{DFC}$, in conjunction with $Q$, are fed into a two-layer bidirectional cross-attention module. Diverging from the conventional Transformer Decoder architecture, we herein reverse the sequence of token self-attention and cross-attention operations. As highlighted in \cite{cheng2022masked}, tokens are initially randomly initialized and devoid of image feature information, rendering self-attention operations ineffective for acquiring sufficient information at the initial stage. Thus, the token-to-feature cross-attention operation is performed. $Q$ serves as the Query, with Keys and Values derived from $F_{DFC}$, to align $Q$ with image features. The output then undergoes residual connection and normalization processing. The formula is as follows.
\begin{equation}
    \label{deqn_ex4}
    Q^\prime=LN(Q+CA(Q,F_{DFC}))
\end{equation}

Subsequently, $Q^\prime$ undergoes self-attention processing to model long-range dependencies and global context among tokens. This is followed by residual normalization and fully connected processing to maintain training stability. Next, a feature-to-token cross-attention operation is performed, where the previous $F_{DFC}$ serves as the Query, the self-attention processed token acts as both Key and Value. This allows $F_{DFC}$ to receive global semantic supplementation from the tokens, facilitating the understanding of the overall structure by $F_{DFC}$ features. The resulting output $F_{DFC}^\prime$ is directed into two branches: feature branch and token branch. In the feature branch, to incorporate spatial details and address the issue of blurred segmentation boundaries, $F_{DFC}^\prime$ undergoes shape transformation and transposed convolution operations to align their resolution with $F_{multiscale}$ and $F_{HQ}$. These three features then undergo additive fusion to supplement the detailed information of the image. The above operations are abstracted as shown in the formula.
\begin{equation}
    \label{deqn_ex5}
    \left\{
    \begin{aligned}
        F_{DFC}^\prime &= CA(F_{DFC},Q^\prime) \\
        F_{final} &= F_{multiscale}+F_{HQ}+TC(F_{DFC}^\prime)
    \end{aligned}
    \right.
\end{equation}

\noindent where $TC(\cdot)$ represents the transposed convolution operations. In the token branch, a token-to-feature cross-attention operation is reintroduced, where $Q^\prime$ serves as the Query and $F_{DFC}^\prime$ acts as both Key and Value. The output undergoes residual normalization and MLP processing, with the tokens ultimately learning the feature importance weights. The formula is represented as follows.
\begin{equation}
    \label{deqn_ex6}
    Q_{final}=LN(Q^\prime+CA(Q^\prime,F_{DFC}^\prime))
\end{equation}

Ultimately, a dynamic convolution mechanism is employed to generate the segmentation mask $M_{seg}$. Specially, convolution kernels and biases are dynamically generated based on the input $Q_{final}$, and these adaptive parameters are then used to perform convolution operations on $F_{final}$. In this way, the convolution weights vary with the input content, enhancing the adaptability to different semantics. The relevant formulas are as follows.
\begin{equation}
    \label{deqn_ex7}
    \left\{
    \begin{aligned}
        \left[w,b\right] &= MLP(Q_{final}) \\
        M_{seg} &= w\ast F_{final}+b
    \end{aligned}
    \right.
\end{equation}

\noindent where $MLP(\cdot)$ denotes a linear mapping module, where $w$ and $b$ represent the generated convolution kernel weights and biases, respectively, and $\ast$ denotes 2D convolution.

\subsection{Probability Map Modulation}

\begin{figure}[!t]
    \centering
    \includegraphics[scale=0.175]{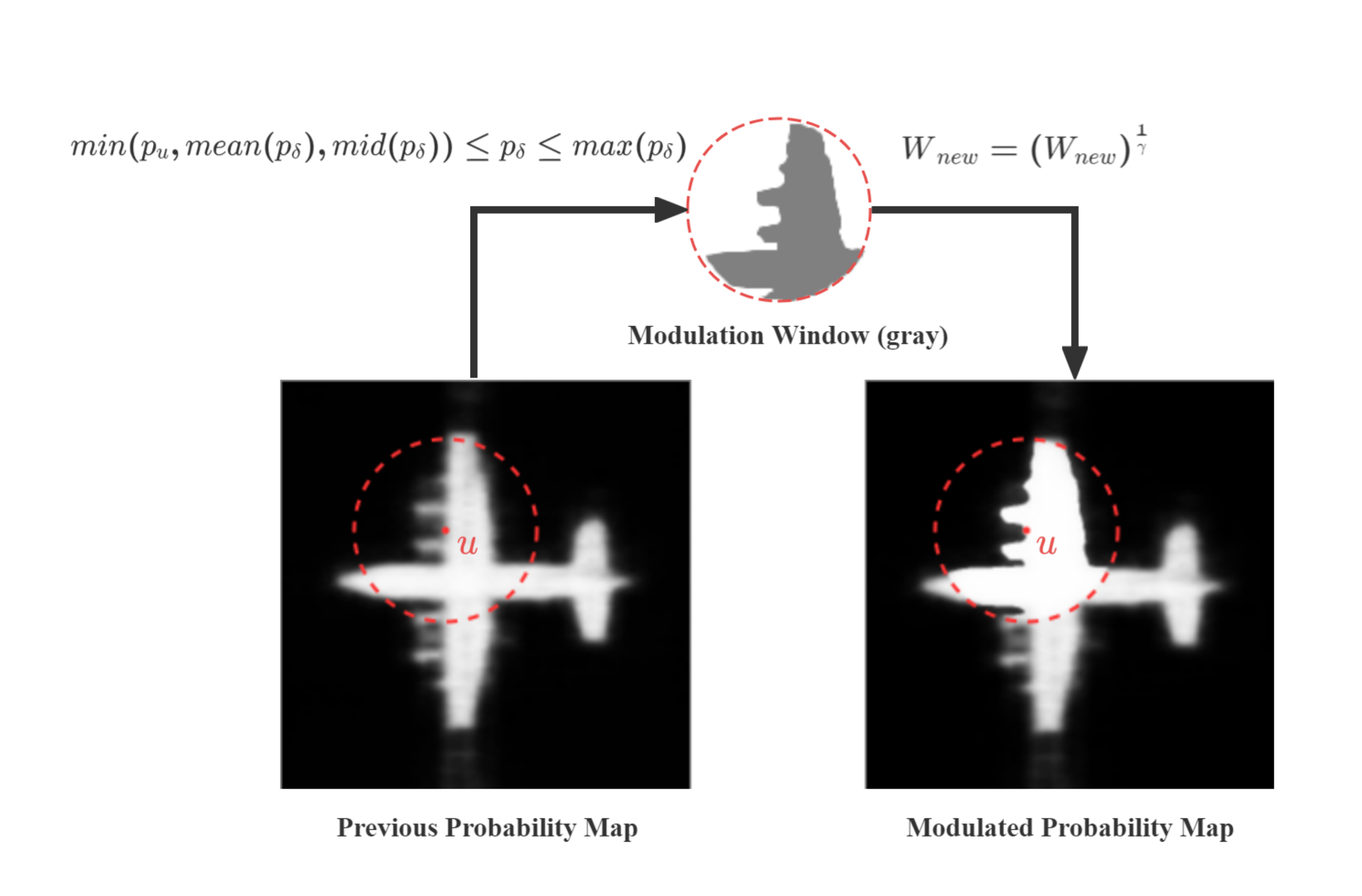}
    \caption{\textbf{Modulate Map Mechanism.} The left image shows the uncorrected result, while the right image presents the corrected version. The modulation window (gray region in the middle image) is extracted using the formula on the left, and the pixel values within this window are refined according to the formula on the right.}
    \label{modulate_map}
\end{figure}

As noted in \cite{lee2024mfp}, probability map modulation enhances or suppresses the probability values of local regions based on the probability map obtained from the previous round and the current user's click information, making them closer to the foreground or background distribution desired by the user. However, in \cite{lee2024mfp}, the modulated local region is a circular area, where both foreground and background pixels may coexist. During the probability modification process, a positive click is intended to enhance the foreground probability, but since there is no distinction between foreground and background within this circular region, the background probability is also inadvertently enhanced. To address this issue, we propose an improved probability map modulation algorithm, as illustrated in Fig.\ref{modulate_map}. 

After the model outputs the predicted probability map, we obtain the latest click point $u=(x,y,t)$ in the current iteration, where $x$ and $y$ represent the horizontal and vertical coordinates, respectively, and $t$ denotes the click type, with $t=1$ indicating a foreground click and $t=0$ indicating a background click. The modulation radius $R$ is then determined based on the click position: if an opposite-type click exists in the vicinity, $R$ is set to half of the distance from the current click to the nearest opposite-type click; otherwise, a predefined radius $R_{max}$ is used. This design prevents the modulation regions of foreground and background from overlapping. To avoid an excessively small modulation radius, we also introduce a minimum radius constraint $R_{min}$. The equation is given as follows.
\begin{equation}
    \label{deqn_ex8}
    R =\left\{
    \begin{aligned}
        &\frac{1}{2}\cdot{min}_{c\in \Omega}\left\|u-c\right\|,     &    \Omega &\neq\emptyset \\
        &R_{max},    &    \Omega &=\emptyset \\
        &R_{min}    &    R &<R_{min} \\
    \end{aligned}
    \right.
\end{equation}

\noindent where $\Omega$ denotes the set of clicks of opposite type, $\left\|\cdot\right\|$ denotes the Euclidean norm. After obtaining the modulation radius, a circular modulation window is constructed with the click point as the center. Considering that the foreground and background pixel probabilities in the probability map exhibit inherent differences, our method introduces conditional filtering to sieve out irrelevant pixels, extracting only foreground or background regions as the modulation window, which differs from the approach in MFP\cite{lee2024mfp} that modulates the entire circular area. The conditional filtering formula is (9).

\begin{figure*}[!t]  
\centering  
\begin{equation}
    \label{deqn_ex9}
    \left\{
    \begin{aligned}
        W &= \left\{ \delta \mid \left\|\delta - u\right\| \leq R \right\},  \\
        W_{new} &= \left\{ \delta \mid min(p_{\delta}) \leq p_{\delta} \leq max(p_u, mean(p_{\delta}), mid(p_{\delta})) \right\}, \quad \quad \quad t_u=0   \\
        W_{new} &= \left\{ \delta \mid min(p_u, mean(p_{\delta}), mid(p_{\delta})) \leq p_{\delta} \leq max(p_{\delta}) \right\}, \quad \quad \quad t_u=1  \\
    \end{aligned}
    \right.
\end{equation}
\end{figure*}

In this formula, $\delta$ represents a specific pixel point in the image, $p_{\delta}$ $(\delta \in W)$ represents the probability corresponding to pixel point $\delta$, where $\delta$ is located within the circular modulation region $W$, $mean(\cdot)$ represents the average value of the probability pixels in the region, $mid(\cdot)$ refers to the median of the probability pixels in the region, and $p_u$ denotes the pixel probability at the click point. Depending on the click type $t_u$, pixels whose probability values fall between the minimum and maximum screening thresholds are retained. Subsequently, the $\gamma$ modulation follows the similar approach of MFP\cite{lee2024mfp}: for different click types, the pixel probabilities in the new modulation region are adjusted, with the adjustment magnitude decreasing as the distance from the click point increases. The relevant formula is (10).

\begin{figure*}[!t]
\centering
\begin{equation}
    \label{deqn_ex10}
    \left\{
    \begin{aligned}
        When &\quad t_u=0: \\
        &\gamma_{max}=\log_{p_u}{(0.01)},   \\
        &\gamma_\delta=\gamma_{max}\cdot\left(1-\frac{d_\delta}{R}\right)+\frac{d_\delta}{R},  \\
        &{\widetilde{W}}_{new}^\delta=\left(W_{new}^\delta\right)^{\gamma_\delta}  \\
        When &\quad t_u=1: \\
        &\gamma_{max}=\frac{1}{\log_{p_u}{\left(0.99\right)}},   \\
        &\gamma_\delta=\gamma_{max}\cdot\left(1-\frac{d_\delta}{R}\right)+\frac{d_\delta}{R},   \\
        &{\widetilde{W}}_{new}^\delta=\left(W_{new}^\delta\right)^\frac{1}{\gamma_\delta}  \\
    \end{aligned}
    \right.
\end{equation}
\end{figure*}

In the formula, $d_\delta=\left\|\delta-u\right\|$ represents the distance between the pixel $\delta$ and the click $u$. The modulated probability map will serve as the input for the next iteration of the model. By leveraging the improved probability map modulation algorithm, segmentation errors in target boundary regions can be corrected, thereby enhancing the accuracy and efficiency of interactive segmentation.


\section{Experiments}
\subsection{Datasets}
To enable the model to effectively accomplish the interactive segmentation task, we selected six high-resolution remote sensing datasets, which were divided into the training set and test set, respectively. Since most of these datasets are originally designed for semantic segmentation tasks, we treated the mask of each closed region in the annotation files as an instance, similar to the datasets used in traditional interactive segmentation tasks.

\textbf{iSAID}\cite{waqas2019isaid} is the first benchmark dataset for instance segmentation in aerial imagery. It contains 655,451 object instances across 15 categories in 2,806 high-resolution images collected from multiple sensors and platforms. These images are mostly optical (RGB three-bands) with a few panchromatic ones, having a resolution ranging from 0.5-5m and widths from 800-13,000 pixels. Precise per-pixel annotations for each instance are provided, ensuring accurate localization for detailed scene analysis.

\textbf{ISPRS Potsdam}\cite{rottensteiner2014isprs} is a benchmark dataset for remote sensing image semantic segmentation, especially for urban area interpretation. It consists of 38 images with a resolution of 0.05m, all having a width of 6000 pixels and contains near-infrared and RGB bands. The dataset contains six categories: buildings, vehicles as instance categories, and trees, impervious surfaces, low vegetation and background as non-instance categories.

\textbf{LoveDA Urban}\cite{wang2021loveda} is the urban scene segment of the LoveDA dataset. Released by the team from Wuhan University, this dataset contains 1,833 high-resolution optical remote sensing images, all collected in July 2016 from three cities: Nanjing, Changzhou and Wuhan, with the data sourced from Google Earth. The images measure 1024×1024 pixels, provide three RGB channels, and have a ground sampling distance of 30 centimeters. It covers seven land cover categories, including background, buildings and roads. Except for the buildings category, no other categories can be used as instances for experiments. Therefore, in this experiment, buildings are divided into individual instances.

\textbf{SandBar} is a remote sensing instance dataset focus on river sandbars. It is primarily derived from LandSat-8 and GF-2 remote sensing images of the upper reaches of the Yellow River in China, documenting the conditions of mid-river sandbars in this region. Images in the dataset are three-band optical images with a resolution of 512×512. All annotations and file archiving follow the COCO format. In the experiments, sandbars will be identified as independent instances.

\textbf{NWPU}\cite{cheng2014multi} is an object detection dataset that has been augmented with an instance segmentation version. It is sourced from aerial sensors and primarily consists of optical images (RGB three-band) with a few panchromatic images. The dataset features a resolution range of 0.08-2m across 10 categories, all of which are considered instance categories. In this study, the positive image set is utilized, comprising 650 images with widths varying from 500 to 2000 pixels. The categories are all considered as instance categories in the experiments.

\textbf{WHUBuilding}\cite{ji2018fully} is a remote sensing image dataset specifically designed for building-related applications. It comprises two main subsets: Satellite Dataset I and Satellite Dataset II. Satellite Dataset I includes 204 high-resolution satellite images captured by sensors such as QuickBird and WorldView, covering various cities globally, with resolutions ranging from 0.3 to 2.5 meters. Satellite Dataset II focuses on East Asia, consisting of 17,388 adjacent satellite images with a uniform resolution of 0.45 meters and a width of 512 pixels, covering 25,749 buildings. In the experiments, one building is considered an instance.

In this experiment, iSAID and ISPRS Potsdam are adopted as the training datasets, encompassing the majority of instance objects in high-resolution images. During the training phase, images from iSAID and ISPRS Potsdam were cropped to a size of 800×800 pixels, with patches lacking annotation masks excluded. Additionally, the ISPRS Potsdam uses RGB bands. The test datasets employed in the experiments consist of iSAID Val, SandBar, LoveDA Urban, NWPU and WHUBuilding, which differ from the training datasets in terms of imaging conditions, resolution and spectral characteristics.

\begin{table*}[t!] 
    \centering
    \vspace{-5pt} 
    \caption{Comparison of our method with state-of-the-art IIS approaches in NoC80, NoC85 and NoC90 on the iSAID, SandBar, NWPU, LoveDA Urban and WHUBuilding datasets. The best results are highlighted in bold, while the second-best are underlined. * denotes that the backbone parameters of SAM in SAM-HQ are frozen. ** indicates that PiClick was trained without multi-output prediction, using the same model and loss function as the original PiClick.} 
    \label{NoC_Contrast}
    \begin{adjustbox}{width=\textwidth}
        \begin{tabular}{l l c c c c c c c c c c c c c c c}
            \toprule
            \multirow{2}{*}{Method} & \multirow{2}{*}{Backbone} & \multicolumn{3}{c}{iSAID} & \multicolumn{3}{c}{SandBar} & \multicolumn{3}{c}{NWPU} & \multicolumn{3}{c}{LoveDA Urban} & \multicolumn{3}{c}{WHUBuilding} \\
            \cmidrule(lr){3-5} \cmidrule(lr){6-8} \cmidrule(lr){9-11} \cmidrule(lr){12-14} \cmidrule(lr){15-17} 
            & & \multicolumn{1}{c}{NoC80} & \multicolumn{1}{c}{NoC85} & \multicolumn{1}{c}{NoC90} & \multicolumn{1}{c}{NoC80} & \multicolumn{1}{c}{NoC85} & \multicolumn{1}{c}{NoC90} & \multicolumn{1}{c}{NoC80} & \multicolumn{1}{c}{NoC85} & \multicolumn{1}{c}{NoC90} & \multicolumn{1}{c}{NoC80} & \multicolumn{1}{c}{NoC85} & \multicolumn{1}{c}{NoC90} & \multicolumn{1}{c}{NoC80} & \multicolumn{1}{c}{NoC85} & \multicolumn{1}{c}{NoC90} \\
            \midrule
            RITM & HRNet32 & 2.37 & 3.74 & 6.71 & 5.11 & 8.29 & 13.54 & 3.25 & \underline{4.51} & 7.17 & 6.73 & 8.36 & 10.82 & 3.82 & 5.41 & 9.00 \\
            CDNet & ResNet34 & 5.47 & 8.14 & 12.09 & 8.45 & 11.76 & 16.06 & 5.30 & 7.84 & 10.57 & 9.73 & 12.58 & 16.17 & 7.36 & 10.20 & 14.54 \\
            FocalClick & HRNet32 & 6.14 & 8.28 & 12.14 & 8.78 & 11.85 & 15.29 & 5.74 & 7.78 & 10.70 & 9.24 & 11.84 & 15.14 & 7.51 & 10.06 & 14.24 \\
            GPCIS & ResNet50 & 3.78 & 5.84 & 9.66 & 8.35 & 11.34 & 15.09 & 3.87 & 5.67 & 8.77 & 7.00 & 9.18 & 12.51 & 5.48 & 7.65 & 11.77 \\
            SAM-HQ* & ViT-B  & 10.70 & 13.47 & 16.71 & 11.14 & 14.98 & 18.29 & 11.63 & 13.31 & 15.31 & 15.95 & 17.71 & 19.15 & 11.23 & 14.40 & 18.13 \\
            SimpleClick & ViT-B & 2.40 & 3.76 & 6.66 & 5.25 & 8.22 & 13.34 & \underline{3.08} & 4.57 & \underline{7.12} & \underline{6.82} & \underline{8.33} & \underline{10.62} & \underline{3.64} & \underline{5.08} & \textbf{8.42} \\
            PiClick** & ViT-B & 3.01 & 4.61 & 8.28 & 5.42 & 8.64 & 14.06 & 4.12 & 6.06 & 8.61 & 8.31 & 10.16 & 12.85 & 3.87 & 5.35 & 8.98 \\
            AdaptiveClick & ViT-B & 2.89 & 4.61 & 7.79 & 6.29 & 9.37 & 13.76 & 3.72 & 5.36 & 7.81 & 6.84 & 8.44 & 10.67 & \textbf{3.54} & \textbf{5.03} & \underline{8.46} \\
            MST & ViT-B & 2.77 & 4.40 & 8.20 & 6.15 & 9.55 & 14.19 & \underline{3.08} & 4.75 & 7.71 & 6.84 & 8.72 & 11.60 & 4.16 & 6.17 & 10.55 \\
            MFP & ViT-B & \textbf{2.20} & \underline{3.46} & \underline{6.62} & \underline{4.63} & \underline{7.91} & \underline{12.97} & 3.24 & 4.56 & 7.30 & 7.13 & 8.79 & 11.62 & 4.01 & 5.63 & 9.28 \\
            \midrule
            $\text{VFM-ISRefiner}_{\text{(ours)}}$ & ViT-B & \underline{2.25} & \textbf{3.38} & \textbf{6.51} & \textbf{4.54} & \textbf{7.67} & \textbf{12.90} & \textbf{2.53} & \textbf{3.83} & \textbf{6.56} & \textbf{5.99} & \textbf{7.62} & \textbf{10.19} & 3.65 & \textbf{5.03} & \textbf{8.42} \\
            \bottomrule 
        \end{tabular}
    \end{adjustbox}
\end{table*}

\begin{table*}[t!] 
    \centering
    \vspace{-5pt} 
    \caption{Comparison of our method against other fine-tuning approaches in NoC85 and NoC90 on the iSAID, SandBar, NWPU, LoveDA Urban and WHUBuilding datasets. The best results are shown in bold. * indicates that the rank of LoRA fine-tuning is 16} 
    \label{Fine_Tune_Contrast}
    \begin{adjustbox}{width=\textwidth}
        \begin{tabular}{@{}l cc cc cc cc cc@{}}
            \toprule
                \multirow{2}{*}{} & \multicolumn{2}{c}{iSAID} & \multicolumn{2}{c}{SandBar} & \multicolumn{2}{c}{NWPU} & \multicolumn{2}{c}{LoveDA Urban} & \multicolumn{2}{c}{WHUBuilding} \\
                \cmidrule(lr){2-3} \cmidrule(lr){4-5} \cmidrule(lr){6-7} \cmidrule(lr){8-9} \cmidrule(lr){10-11}
                & NoC85 & NoC90 & NoC85 & NoC90 & NoC85 & NoC90 & NoC85 & NoC90 & NoC85 & NoC90 \\  
                \midrule
                Pretrained Weights & 5.40 & 8.51 & 8.69 & 13.24 & 4.91 & 7.39 & 7.66 & 10.46 & 6.38 & 9.97 \\
                Full Fine-Tuning & 3.46 & 6.62 & 7.91 & 12.97 & 4.56 & 7.30 & 8.79 & 11.62 & 5.63 & 9.28 \\
                LoRA Fine-Tuning * & 3.52 & 6.64 & 8.14 & 13.33 & 4.73 & 7.41 & 8.42 & 11.18 & 5.23 & 8.74\\
                \midrule
                $\text{VFM-ISRefiner}_{\text{(ours)}}$ & \textbf{3.38} & \textbf{6.51} & \textbf{7.67} & \textbf{12.90} & \textbf{3.83} & \textbf{6.56} & \textbf{7.62} & \textbf{10.19} & \textbf{5.03} & \textbf{8.42}\\
            \bottomrule 
        \end{tabular}
    \end{adjustbox}
\end{table*}

\subsection{Implementation Details}

\textbf{1) Model Details.} Since our work is based on the adapter set up for the MFP\cite{lee2024mfp} model, the parameters of the frozen part adopt the model parameters of MFP\cite{lee2024mfp} trained on the COCOLVIS\cite{sofiiuk2022reviving} dataset. The backbone is ViT-Base, with the patch embedding size input to the backbone being (16, 16) and the feature dimension being 768. The output of the backbone is processed by SimpleFPN, and the multi-scale feature dimensions of the output are (128, 256, 512, 1024). The Image Feature Extractor consists of one residual block and three deformable residual blocks, with an input dimension of 3 and an output dimension of 768. In the Image Feature Fusion module, there are two cross-attention layers with 8 heads where both the input and output dimensions are 768. Subsequently, the features undergo an upsampling operation to quadruple the original resolution, resulting in an output dimension of 192. In the Transformer Decoder components, there are two decoder layers with input dimension of 768. Each decoder layer contains two cross-attention modules and one self-attention module, with each attention module having 8 heads. In the dynamic convolution fusion stage of token and feature map, the convolution kernel size is $3\times3$, with an input dimension of 192 and an output dimension of 1.

\begin{figure*}[!t]
    \centering
    \includegraphics[scale=0.235]{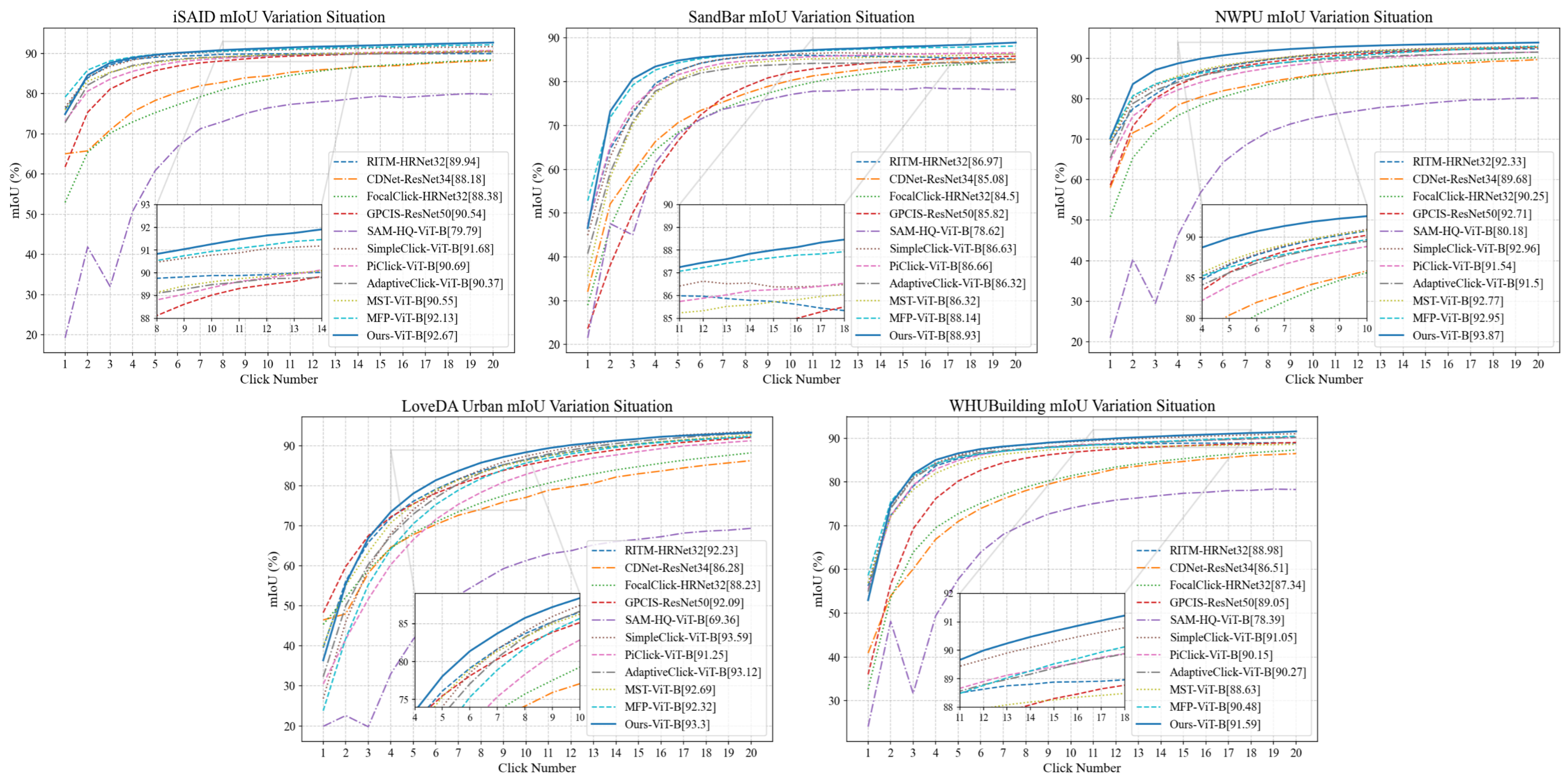}
    \caption{\textbf{mIoU comparison} of our method and other approaches on the iSAID, SandBar, NWPU, LoveDA Urban and WHUBuilding datasets. The horizontal axis denotes the number of clicks, while the vertical axis represents the mean IoU. In the legend, "RITM-HRNET32[89.94]" denotes the RITM model with HRNET32 backbone, which achieves a maximum mIoU value of 89.94\% on the dataset. }
    \label{MIoU_Analysis}
\end{figure*}

\textbf{2) Training Settings.} To generate training data, we combine the iSAID dataset and the ISPRS Potsdam dataset. Images from both datasets are randomly cropped to $448\times448$. In addition, we perform a series of image augmentation operations, such as random rotation, flipping, brightness adjustment and RGB modification. Regarding the strategies for click points generation and sampling, we adopt the iterative learning framework proposed in RITM\cite{sofiiuk2022reviving} and utilize the training sample generation approach introduced in MFP\cite{lee2024mfp} to sample positive and negative clicks. The maximum number of click points per sample is set to 24.

During the training process, we apply the Adam optimizer to minimize the normalized focal loss with $\beta_1=0.9$ and $\beta_2=0.999$. The model is trained for 60 epochs, with the learning rate initially set to $5\times{10}^{-5}$ and reduced by a factor of 10 at the 10th and 50th epoch. Each epoch comprises 30000 training samples. All models are trained on 2 NVIDIA RTX 4090 GPUs, with the batch size configured to 16. Additionally, we fix the value of the hyperparameter associated with probability map modulation to $R_{max}=100$, as set in MFP\cite{lee2024mfp}.

\textbf{3) Evaluation Metrics.} Adhering to the evaluation strategies proposed by previous works, we adopt average number of clicks (NoC) required to achieve a certain Intersection-over-Union (IoU) ratio. The maximum number of annotated clicks is set to 20. To evaluate performance in high-precision segmentation scenarios, we set the target IoU ratio to 90\% and report NoC80, NoC85 and NoC90. Additionally, we plot the mean Intersection-over-Union (mIoU) scores against the number of clicks.

\begin{figure*}[!t]
  \centering
  \includegraphics[scale=0.28]{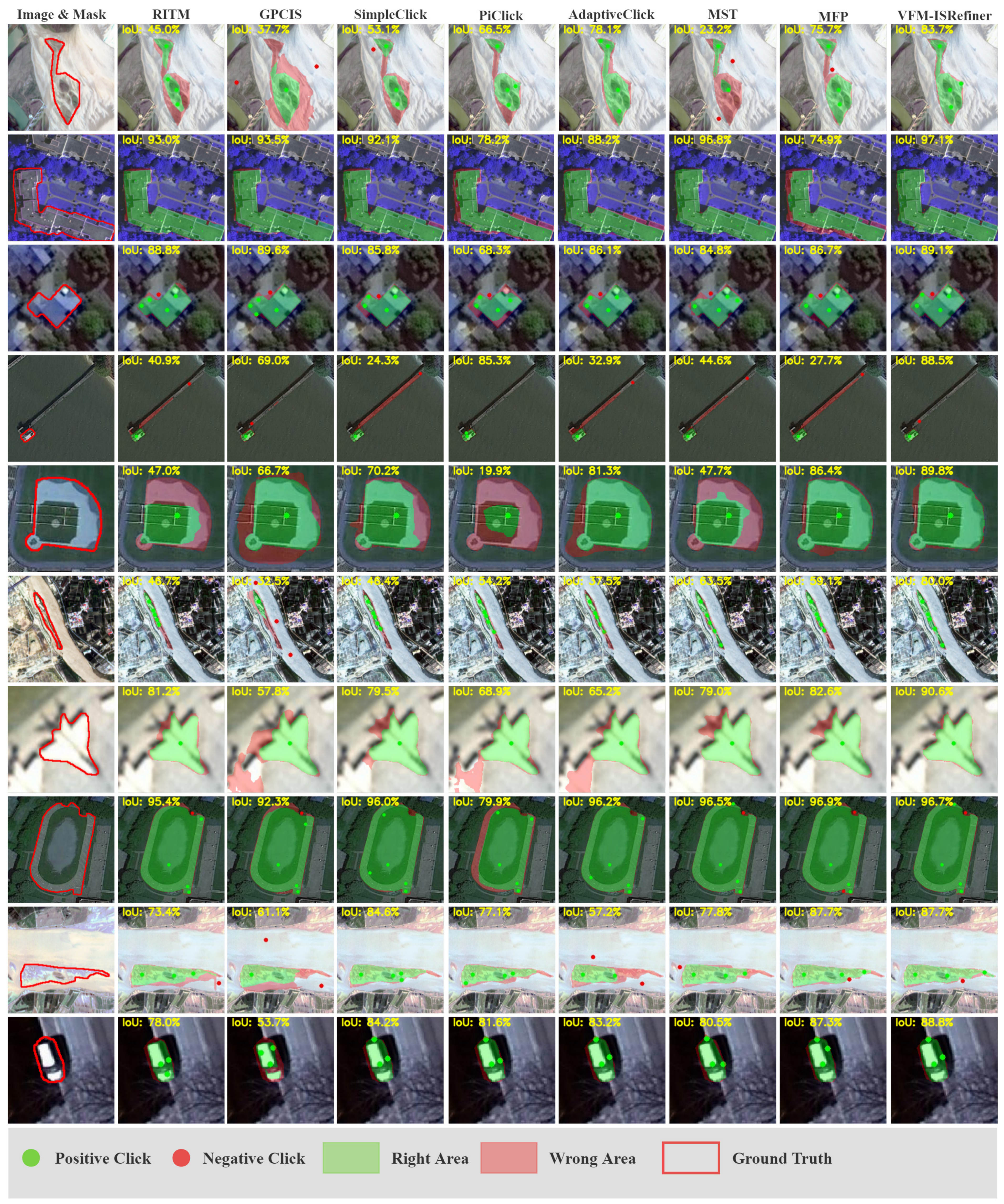}
  \caption{\textbf{Segmentation results} of RITM(HRNet32), GPCIS(ResNet50), SimpleClick(ViT-B), PiClick(ViT-B), AdaptiveClick(ViT-B), MST(ViT-B), MFP(ViT-B) and our VFM-ISRefiner(ViT-B). Red clicks indicate negative clicks, and green clicks indicate positive clicks. Red regions represent mismatches between the predicted mask and the ground truth mask, while green regions represent matched areas. The red bounding box denotes the ground truth mask boundary.}
  \label{visual_results}
\end{figure*}

\subsection{Comparative Experiment}

In this subsection, we select 10 state-of-the-art interactive segmentation algorithms for comparison with the algorithm proposed in this paper. Among them, RITM\cite{sofiiuk2022reviving}, CDNet\cite{chen2021conditional}, FocalClick\cite{chen2022focalclick} and GPCIS\cite{zhou2023interactive} adopt convolutional neural networks as their feature extraction backbones. RITM\cite{sofiiuk2022reviving} and FocalClick\cite{chen2022focalclick} employ HRNet32 as their backbones, while CDNet\cite{chen2021conditional} and GPCIS\cite{zhou2023interactive} utilize ResNet34 and ResNet50 as their backbones, respectively; SAM-HQ\cite{ke2023segment} uses a frozen SAM\cite{kirillov2023segment} model as its backbone, where the SAM\cite{kirillov2023segment} backbone is pretrained on large-scale image datasets. Leveraging the strong feature generalization capability of the backbone, SAM-HQ\cite{ke2023segment} incorporates an adaptation module to achieve better segmentation performance on specific tasks. SimpleClick\cite{liu2023simpleclick}, PiClick\cite{yan2023piclick}, AdaptiveClick\cite{lin2024adaptiveclick}, MST\cite{xu2025mst} and MFP\cite{lee2024mfp} all use ViT-B as their feature extraction backbones, with PiClick\cite{yan2023piclick} and AdaptiveClick\cite{lin2024adaptiveclick} following a DETR-like architecture.

All the above 10 models are trained on the remote sensing training data mentioned before. The frozen backbone of our model is not fine-tuned on remote sensing images; instead, the model adapter is trained directly using the model parameters pretrained on natural images. Correspondingly, the SAM backbone used in SAM-HQ\cite{ke2023segment} is also not fine-tuned for the remote sensing dataset, and the parameters of SAM\cite{kirillov2023segment} backbone are frozen throughout the training process. Apart from this, the other nine models undergo full-parameter training. Additionally, PiClick\cite{yan2023piclick} necessitates hierarchical parent-child relationships among dataset categories to enable the generation of multi-label outputs. However, given the absence of such categorical hierarchies in public remote sensing datasets, we opt to train PiClick\cite{yan2023piclick} using a variant that abstains from producing multiple annotation results.

In addition, this subsection will explore the impact of different fine-tuning methods on model performance, comparing the segmentation effectiveness and model efficiency among the pretrained model, the fully parameter-trained model, the LoRA fine-tuning model, and our proposed model. Notably, the backbone parameters are frozen during the LoRA fine-tuning process.

\textbf{1) Comparison of NoC.} As shown in Table \ref{NoC_Contrast}, our VFM-ISRefiner method exhibits strong competitiveness and stability overall across multiple remote sensing image segmentation datasets such as iSAID, SandBar, NWPU, LoveDA Urban and WHUBuilding, as well as under different intersection over union (IoU) thresholds including NoC80, NoC85 and NoC90. At lower NoC80 threshold, our method achieves optimal performance on the SandBar, NWPU and LoveDA Urban datasets, and suboptimal performance on the iSAID dataset. However, at the higher IoU thresholds of NoC85 and NoC90, our method attains the best results on all five datasets. Specifically, the average number of clicks required to reach the IoU 85\% and IoU 90\% thresholds is reduced by 0.352 and 0.242 respectively compared to that of the suboptimal method. This highlights the distinct advantage of our method in refined interactive segmentation, making it suitable for high-quality segmentation and annotation of instance objects in remote sensing images.

It is noteworthy that SAM-HQ\cite{ke2023segment}, which adopts a similar technical approach to our method, exhibits the poorest performance among all models in terms of NoC metric. We attribute this to the fact that the backbone of SAM-HQ\cite{ke2023segment} is pretrained on natural images. Due to the differences between remote sensing images and natural images in aspects such as radiometric energy and spectral accuracy, its strong generalization capability fails to be effectively manifested in remote sensing scenarios. Furthermore, the simplistic design of the adapter module in SAM-HQ\cite{ke2023segment} renders it incapable of adequately learning and extracting relevant features from remote sensing images, thereby resulting in suboptimal performance. In contrast, the adapter module of our method integrates convolutional networks with the global attention mechanism of Transformer, which can compensate for the deficiency of the backbone in extracting features specific to remote sensing images. In addition, the improved probability map modulation method proposed in this paper can correct the segmentation errors of target boundaries, thereby further improving the segmentation accuracy of our model.

\begin{table}[htbp]
  \centering
  \caption{Efficiency Comparison of Different IIS Methods. $^\dagger$ denotes testing time per interaction}
  \label{efficiency_comparison_diff_method}
  \setlength{\tabcolsep}{3.5pt}
  \begin{tabular}{lcccc}
      \toprule
      Method     & Backbone & Trainable Params. & FLOPs  & Time $^\dagger$ \\
      \midrule
      RITM        & HRNet32    & 30.95M     & 103.74G   & 65ms           \\
      SAM-HQ      & ViT-B    & 2.09M     & 55.04G   & 50.8ms        \\
      SimpleClick & ViT-B    & 98.03M    & 169.78G   & 29.6ms           \\
      PiClick     & ViT-B    & 116.28M    & 197.40G   & 87ms           \\
      AdaptiveClick     & ViT-B    & 121.89M    & 269.81G   & 70.6ms           \\
      MST     & ViT-B    & 155.52M    & 239.52G   & 77.8ms           \\
      MFP     & ViT-B    & 98.66M    & 181.3G   & 34.2ms           \\
      \midrule
      $\text{VFM-ISRefiner}_{\text{(ours)}}$     & ViT-B    & 40.11M    & 212.49G   & 48.2ms           \\

      \bottomrule
  \end{tabular}
\end{table}

\begin{table}[htbp]
  \centering
  \caption{Efficiency Comparison of Different Fine-Tuning Methods . $^\dagger$ denotes testing time per interaction}
  \label{efficiency_comparison_diff_finetuning}
  \setlength{\tabcolsep}{3.5pt}
  \begin{tabular}{lccc}
      \toprule
      Method  & Trainable Params. & FLOPs  & Time $^\dagger$ \\
      \midrule
      Pretrained Weights    & 0M    & 181.3G    & 35.4ms           \\
      Full Fine-Tuning      & 98.66M     & 181.29G   & 34.2ms        \\
      LoRA Fine-Tuning      & 1.1M    & 181.3G    & 39.8ms           \\
      \midrule
      $\text{VFM-ISRefiner}_{\text{(ours)}}$     & 40.11M    & 212.49G   & 48.2ms           \\
      \bottomrule
  \end{tabular}
\end{table}

\textbf{2) Comparison of mIoU Variation.} As illustrated in Fig.\ref{MIoU_Analysis}, we adopt the number of clicks as the independent variable and mIoU as the evaluation metric to conduct a comparative performance analysis of competing methods; specifically, while the initial mIoU of our proposed method at the first click is not competitive, its mIoU curve exhibits a steep upward trend with increasing click counts, rising sharply within the first 5-10 clicks as reflected in the subfigures corresponding to each dataset, and then rapidly approaches a high-performance plateau that stabilizes after 10-15 clicks. After the final 20th click, our method attains the optimal mIoU among all competing methods on four datasets, namely iSAID (92.67\%), SandBar (88.93\%), NWPU (93.87\%), and WHUBuilding (91.59\%), and it still achieves the second-highest mIoU with a value of 93.3\% on the LoveDA Urban dataset. Except for the SandBar dataset, the final mIoU values of iSAID, NWPU, LoveDA Urban, and WHUBuilding all surpass 90\%, and notably, the iSAID and NWPU datasets, where our method reaches the 90\% mIoU threshold within 8 and 10 clicks, respectively demonstrate that our method is capable of rapidly converging to a high mIoU level with a small number of interaction clicks, a characteristic clearly reflected in the curve trends of these two datasets. Collectively, the experimental results across the five datasets indicate that our method possesses robust adaptability and prominent performance advantages in terms of both rapid convergence and high final accuracy, thereby fully validating the effectiveness and precision of the proposed method in click-based interactive segmentation tasks.

\begin{figure*}[!t]
    \centering
    \includegraphics[scale=0.215]{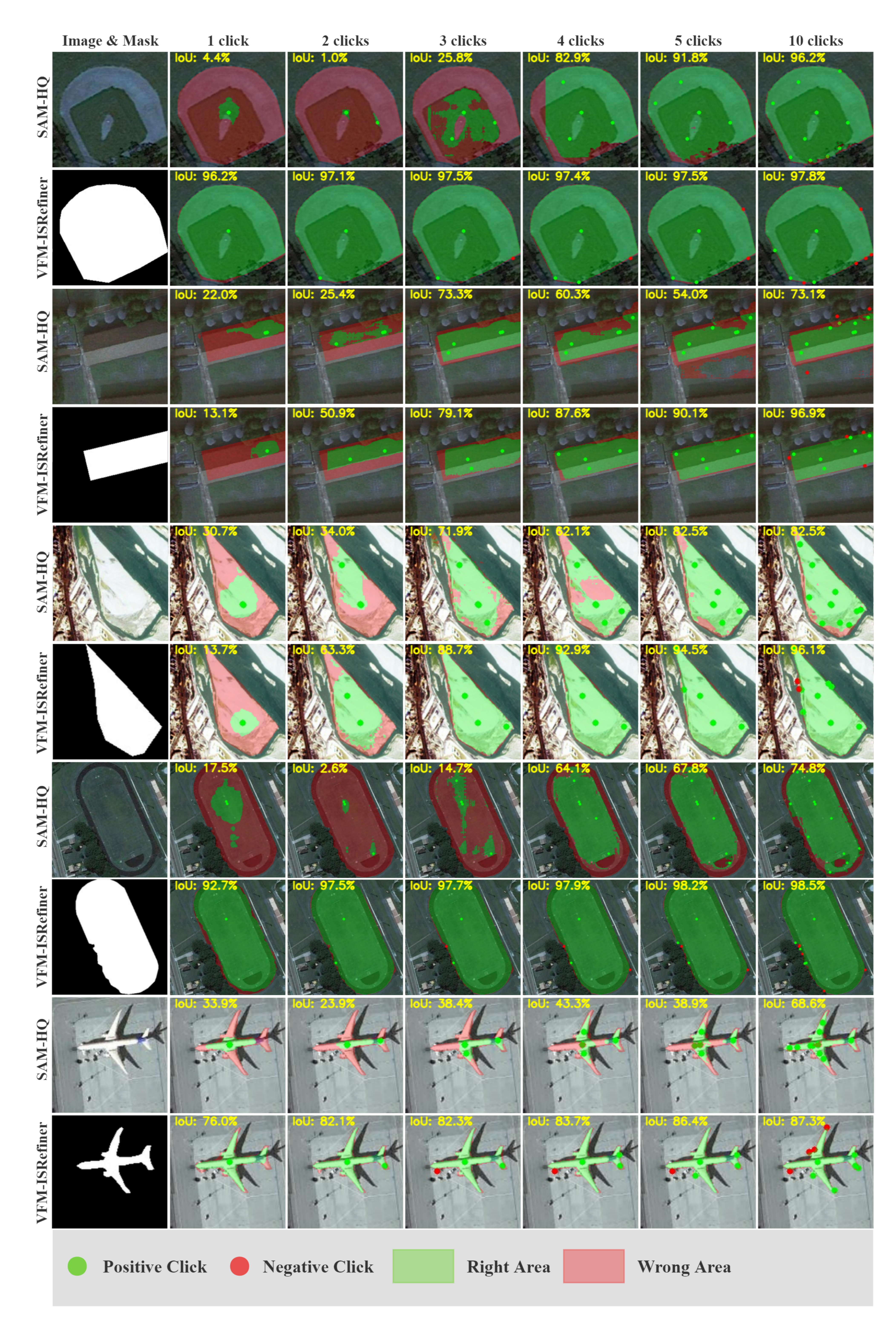}
    \caption{\textbf{Segmentation results} of our method and SAM-HQ, which adopts a mechanism similar to ours. The first column shows the original images and their ground truth masks, while the second to seventh columns illustrate the segmentation results under different numbers of clicks.}
    \label{Multi_Points}
\end{figure*}

\textbf{3) Comparison With Fine-Tune Methods.} As presented in Table \ref{Fine_Tune_Contrast}, the first row presents the experimental results using the MFP weights pretrained on natural images. The second row shows the results of MFP with full-parameter fine-tuning on the remote sensing dataset proposed in this paper. The third row denotes the experimental results of LoRA fine-tuning applied to the ViT backbone of MFP. The fourth row reports the results obtained by the fine-tuning method proposed in this study. Our proposed adapter fine-tuning method achieves the optimal performance across the NoC85 and NoC90 metrics on five remote sensing datasets: iSAID, SandBar, NWPU, LoveDA Urban and WHUBuilding. Compared with the pretrained weights, it reduces the NoC85 score from 5.40 to 3.38 on the iSAID dataset, significantly enhancing the model's adaptability to the remote sensing domain. Against full fine-tuning, it decreases the NoC85 from 7.91 to 7.67 and NoC90 from 12.97 to 12.90 on the SandBar dataset, maintaining higher segmentation accuracy while reducing parameter updates and avoiding the potential overwrite of pretrained general visual knowledge by full fine-tuning. Compared with LoRA fine-tuning, it lowers the NoC85 from 4.73 to 3.83 and NoC90 from 7.41 to 6.56 on the NWPU dataset, overcoming the limitation of LoRA's low-rank structure in expressing remote sensing-specific features (e.g., irregular boundaries, spectral heterogeneity). Ultimately, it balances the preservation of general representations, efficient domain adaptation, and reduced interaction cost, providing an efficient and high-performance fine-tuning solution for interactive remote sensing image segmentation.

\textbf{4) Comparison of Efficiency.} In this subsubsection of comparisons, we have calculated the number of trainable parameters, FLOPs (Floating Point Operations) of the models, as well as the average time consumed per interaction across the five test datasets. It should be noted that when calculated the average consumed time, all models were tested under identical conditions, including input image size, number of test clicks, hardware specifications for test and so on. As shown in Table \ref{efficiency_comparison_diff_method}, due to the frozen backbone, the number of trainable parameters in our model is significantly smaller than that of models with ViT-based Vision Foundation Models as backbones which undergo full-parameter training. Consequently, during the mode training process, our model involves minimal gradient computations and thus achieves faster training speed. From a comprehensive perspective of FLOPs and average time consumption, although our model exhibits high FLOPs and increased complexity, as it incorporates an additional adapter on top of the Vision Foundation Model, in terms of average time consumption, our model performs favorably, with a shorter average testing time compared to models such as RITM\cite{sofiiuk2022reviving}, PiClick\cite{yan2023piclick} and AdaptiveClick\cite{lin2024adaptiveclick}. This indicates that our model also holds certain advantages in the generation speed of segmentation results, facilitating efficient segmentation and annotation of remote sensing images.

By comparing the indicators of different fine-tuning methods in the Table \ref{efficiency_comparison_diff_finetuning}, although our method has slightly higher interactive testing time and computational complexity than other approaches, its trainable parameters (40.11M) are significantly superior to those of Full Fine-Tuning (98.66M). While ensuring a certain degree of parameter update flexibility to adapt to the task, it greatly reduces the parameter scale compared with full-parameter training, achieving a balance between parameter efficiency and task adaptability. This provides a better choice for scenarios that require moderate parameter adjustments to improve performance.

\textbf{5) Visualization Comparison.} As shown in Fig.\ref{visual_results}, existing methods often suffer from boundary shifts and regional missegmentation in complex remote sensing scenarios. Traditional convolution-based approaches (e.g., RITM\cite{sofiiuk2022reviving}, GPCIS\cite{zhou2023interactive}) tend to have over/under-segmentation with complex backgrounds or blurry boundaries, leading to low IoU scores. ViT-based methods (e.g., SimpleClick\cite{liu2023simpleclick}, PiClick\cite{yan2023piclick}, AdaptiveClick\cite{lin2024adaptiveclick}, MST\cite{xu2025mst}) perform better on samples with regular boundaries (Row 8) but still missegment targets with complex shapes (Rows 1, 2, 6, 9) or under illumination/shadow interference (Rows 7, 10). While MFP\cite{lee2024mfp} optimizes boundary segmentation by refining probability maps, it struggles with small (Row 4) and slender targets (Row 6). In contrast, our improved modulation method achieves stable segmentation across all scenarios, with predictions closely aligning with ground truth and significantly fewer errors. Quantitatively, our IoU exceeds 80\% in most cases (even over 90\% for some complex targets), whereas other methods generally range from 60\% to 80\%. This confirms our method has stronger robustness and boundary adherence in interactive remote sensing image segmentation-especially for complex boundaries or severe background interference-effectively reducing errors and ensuring accurate, complete results.

Since both our method and SAM-HQ\cite{ke2023segment} are built upon fine-tuning Vision Foundation Models within the similar technical paradigm, we choose to conduct a direct comparison with SAM-HQ\cite{ke2023segment}, which makes the evaluation more convincing. As shown in Fig.\ref{Multi_Points}, SAM-HQ\cite{ke2023segment} performs poorly with 1-3 clicks, often producing large mis-segmented regions (Rows 1, 3, 4, 5). Though improving with more clicks, it still exhibits noticeable mis-segmentations along complex boundaries and within objects, with final IoU typically 70\%-90\%. In contrast, our method achieves higher accuracy with fewer interactions. It converges stably as clicks increase, maintaining over 90\% IoU across categories. Even for complex targets like airplanes and irregular buildings, our predictions align well with ground truth, with far fewer errors. The above results demonstrate that under a similar technical framework, our method achieves higher segmentation accuracy while requiring fewer user interactions

\subsection{Ablation Study}
To verify the effectiveness and critical of certain components or techniques in the algorithm, we designed four sets of ablation experiments: the role of ViT output features ${F_{ViT}}$ and high-quality features ${F_{HQ}}$, the impact of different values of fusion parameter $\theta$ in Fig.\ref{High_Quality_Feature_Extractor} on the results, the influence of ViT early feature fusion methods on the original mask correction method and the comparison between the original mask modulation method the improved mask modulation method. The experimental conditions remain consistent with those previously used, with the test datasets being iSAID, SandBar and WHUBuilding, and the evaluation metrics being NoC80 and NoC90.

\textbf{1) The Role of ${F_{ViT}}$ and ${F_{HQ}}$.} As can be observed from Table \ref{F_ViT_F_HQ}, the model achieves the poorest performance when both ${F_{ViT}}$ and ${F_{HQ}}$ are absent, indicating that these two features exert a certain influence on the model's performance. Specifically, the output of Vision Foundation Model possesses global semantic properties. Feeding this output into the model decoder helps supplement the semantic information of remote sensing images. ${F_{HQ}}$ generated by the high-quality feature extraction module, incorporates convolutional operations, enabling it to capture local detailed features in images. This facilitates more refined learning of target boundaries and shapes by the model. Additionally, it is evident that the degree of impact on the model differs significantly between the absence of ${F_{ViT}}$ alone and the absence of ${F_{HQ}}$ alone. The absence of ${F_{HQ}}$ substantially degrades the model's performance, whereas the absence of ${F_{ViT}}$ results in only minor performance loss. This can be attributed to the fact that ${F_{ViT}}$ output by the ViT, undergoes processing by SimpleFPN to generate ${F_{multiscale}}$, with ${F_{multiscale}}$ retaining most of the semantic information from ${F_{ViT}}$. Consequently, the absence of ${F_{ViT}}$ has a limited impact on model performance. Conversely, ${F_{HQ}}$ represents an additionally extracted high-quality image feature that holds high value for model's feature learning. The lack of this feature leads to suboptimal segmentation results for remote sensing targets with complex boundaries. 

\begin{table}[t] 
    \centering
    \caption{Comparison in NoC80 and NoC90 to investigate the effect of ${F_{ViT}}$ and ${F_{HQ}}$ on model performance.} 
    \label{F_ViT_F_HQ}
    \begin{tabular}{@{}cc cc cc cc@{}} 
        \toprule
        \multirow{2}{*}{${F_{ViT}}$} & \multirow{2}{*}{${F_{HQ}}$} & \multicolumn{2}{c}{iSAID} & \multicolumn{2}{c}{SandBar} & \multicolumn{2}{c}{WHUBuilding} \\
        \cmidrule(lr){3-4} \cmidrule(lr){5-6} \cmidrule(lr){7-8}
        & & NoC80 & NoC90 & NoC80 & NoC90 & NoC80 & NoC90 \\ 
        \midrule
        $\usym{2613}$ & $\usym{2613}$ & 3.03 & 8.20 & 6.43 & 14.16 & 3.91 & 9.12 \\
        $\checkmark$ & $\usym{2613}$ & 2.79 & 7.58 & 6.12 & 13.95 & 3.77 & 8.90 \\
        $\usym{2613}$ & $\checkmark$ & 2.29 & 6.74 & 4.84 & 13.24 & 3.68 & 8.49 \\
        $\checkmark$ & $\checkmark$ & \textbf{2.25} & \textbf{6.51} & \textbf{4.54} & \textbf{12.90} & \textbf{3.65} & \textbf{8.42} \\
        \bottomrule 
    \end{tabular}
\end{table}

\begin{table}[htbp] 
    \centering
    \caption{Comparison in NoC80 and NoC90 to analyze the impact of $\theta$ on the model performance.} 
    \label{Theta}
    \begin{tabular}{@{}l cc cc cc@{}}  
        \toprule
        \multirow{2}{*}{} & \multicolumn{2}{c}{iSAID} & \multicolumn{2}{c}{SandBar} & \multicolumn{2}{c}{WHUBuilding} \\
        \cmidrule(lr){2-3} \cmidrule(lr){4-5} \cmidrule(lr){6-7}
        & NoC80 & NoC90 & NoC80 & NoC90 & NoC80 & NoC90 \\ 
        \midrule
        $\theta = 1$ & 2.28 & 6.87 & 5.42 & 13.50 & 3.70 & 8.70 \\
        $\theta = 0.5$ & 2.36 & 6.79 & 4.90 & 13.42 & 3.97 & 9.00 \\
        Learnable $\theta$ & \textbf{2.25} & \textbf{6.51} & \textbf{4.54} & \textbf{12.90} & \textbf{3.65} & \textbf{8.42} \\
        \bottomrule 
    \end{tabular}
\end{table}

\textbf{2) The Impact of $\theta$.} Table \ref{Theta} presents the results of ablation experiments on parameter $\theta$. When $\theta$ is set to a fixed value, the overall performance at $\theta=0.5$ outperforms that at $\theta=1$, indicating that $\theta$ can balance the performance contributions of $F_{IFE}$ and $F_{early}$ during cross-attention fusion. Although a fixed $\theta$ is effective for some situations, it struggles to maintain stable performance across different scenarios. In contrast, setting $\theta$ as a learnable parameter yields better results across all datasets and settings. This confirms that the learnable $\theta$ enables the model to adaptively allocate the importance of different feature sources, thereby enhancing overall robustness and generalization capability.

\begin{table}[t] 
    \centering
    \caption{Comparison in NoC80 and NoC90 to analyze the impact of $F_{ViT}^{early}$ fusion on the model performance.} 
    \label{ViT_Early}
    \begin{tabular}{@{}l cc cc cc@{}}  
        \toprule
        \multirow{2}{*}{} & \multicolumn{2}{c}{iSAID} & \multicolumn{2}{c}{SandBar} & \multicolumn{2}{c}{WHUBuilding} \\
        \cmidrule(lr){2-3} \cmidrule(lr){4-5} \cmidrule(lr){6-7}
        & NoC80 & NoC90 & NoC80 & NoC90 & NoC80 & NoC90 \\ 
        \midrule
        only ${F_{ViT}^1}$ & 2.36 & 7.19 & 5.17 & 13.54 & 3.95 & 8.83 \\
        \addlinespace  
        only ${F_{ViT}^2}$ & 2.32 & 6.63 & 4.93 & 13.21 & 3.85 & 9.08 \\
        \addlinespace  
        Fixed Fusion & 2.29 & 6.97 & 5.06 & 13.48 & 3.74 & 8.56 \\
        \addlinespace  
        Gated Fusion & \textbf{2.25} & \textbf{6.51} & \textbf{4.54} & \textbf{12.90} & \textbf{3.65} & \textbf{8.42} \\
        \bottomrule 
    \end{tabular}
\end{table}

\begin{figure}[t]
    \includegraphics[scale=0.26]{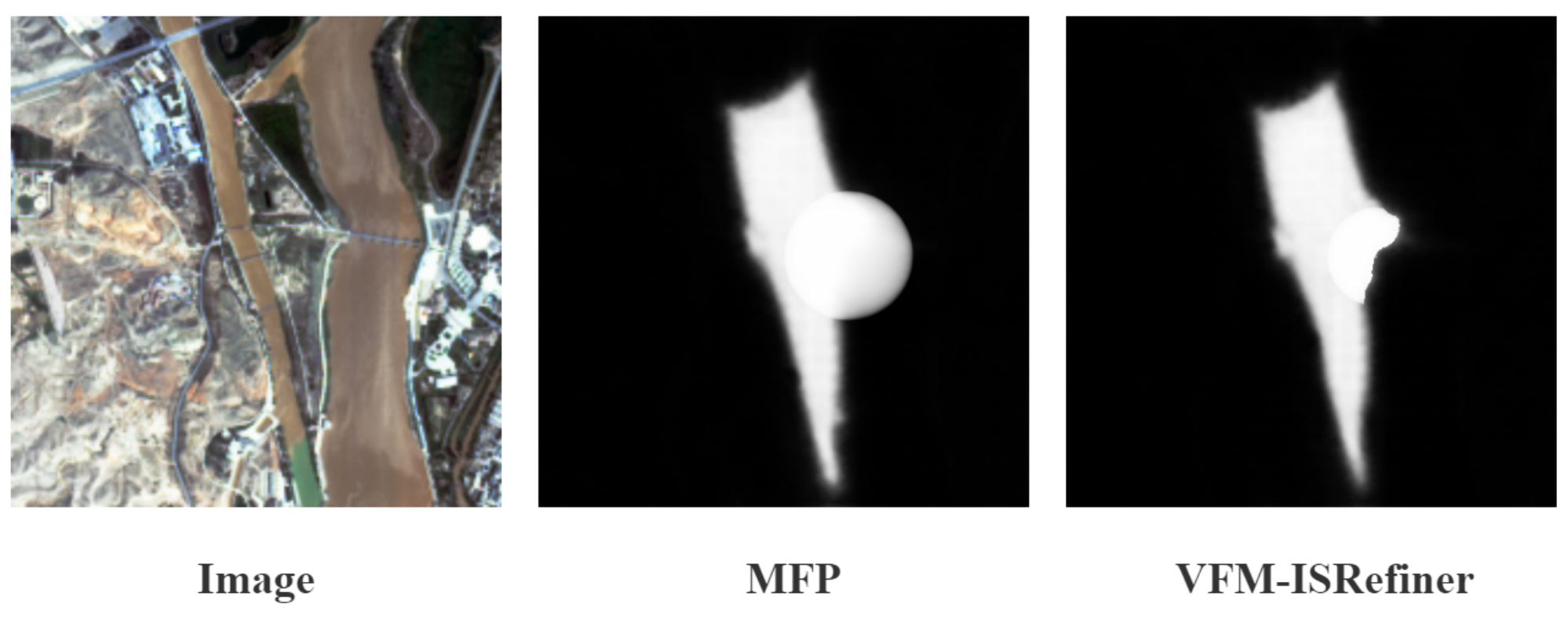}
    \caption{\textbf{Visualization of probability map modulation} using the MFP's algorithm and our proposed algorithm. The left shows the image input, the middle image presents the modulation result of the MFP's algorithm, and the right image illustrates the modulation result of our method.}
    \label{Modulate_Contrast}
\end{figure}

\textbf{3) The Influence of $F_{ViT}^{early}$ Fusion.} Table \ref{ViT_Early} reports the ablation study on different early ViT feature fusion strategies. Using only a single layer feature $F_{ViT}^1$ or $F_{ViT}^2$ leads to suboptimal performance, indicating that features from a single level are insufficient to capture the multi-scale information required in complex scenarios. Fixed Fusion brings moderate improvements, but its inability to adaptively weight different feature sources results in limited gains. In contrast, our proposed Gated Fusion consistently achieves the optimal performance across all datasets and settings. These demonstrate that Gated Fusion can effectively learn the relative importance of features from different layers, exploit their complementarity, and thereby improve the adaptability and reliability of the model in handling diverse and challenging remote sensing conditions.

\begin{table}[htbp] 
    \centering
    \caption{Comparison of the impact of the MFP's modulation algorithm and our proposed algorithm on the model under the NoC80 and NoC90 metrics.} 
    \label{Modulation_Table}
    \begin{tabular}{@{}l cc cc cc@{}}  
        \toprule
        \multirow{2}{*}{} & \multicolumn{2}{c}{iSAID} & \multicolumn{2}{c}{SandBar} & \multicolumn{2}{c}{WHUBuilding} \\
        \cmidrule(lr){2-3} \cmidrule(lr){4-5} \cmidrule(lr){6-7}
        & NoC80 & NoC90 & NoC80 & NoC90 & NoC80 & NoC90 \\  
        \midrule
        MFP & \textbf{2.24} & 6.70 & 4.79 & 13.21 & 3.94 & 9.11 \\
        VFM-ISRefiner & 2.25 & \textbf{6.51} & \textbf{4.54} & \textbf{12.90} & \textbf{3.65} & \textbf{8.42} \\
        \bottomrule 
    \end{tabular}
\end{table}

\textbf{4) The Contrast of MFP's and Our Modulation Method.} To validate the effectiveness of the proposed probability map modulation algorithm, we conduct this ablation study. Quantitatively, as shown in Table \ref{Modulation_Table}, on the iSAID, SandBar and WHUBuilding datasets with NoC80 and NoC90 as metrics, our method consistently reduces the error compared to the baseline. For instance, on the SandBar datasets, the NoC80 metric decreases from 4.79 to 4.54, and the NoC90 metric decreases from 13.21 to 12.90. From a qualitative perspective, Fig.\ref{Modulate_Contrast} illustrates that the baseline method produces spurious "circular" response regions in the modulated probability maps, whereas our approach generates results that more closely adhere to the true object contours. This demonstrates the ability of the proposed algorithm to effectively suppress wrong regions in probability maps. Overall, the improved modulation algorithm provides higher-quality probability maps to support subsequent interactive steps, thereby facilitating both enhanced segmentation accuracy and improved interaction efficiency.

\begin{figure}[t]
    \includegraphics[scale=0.29]{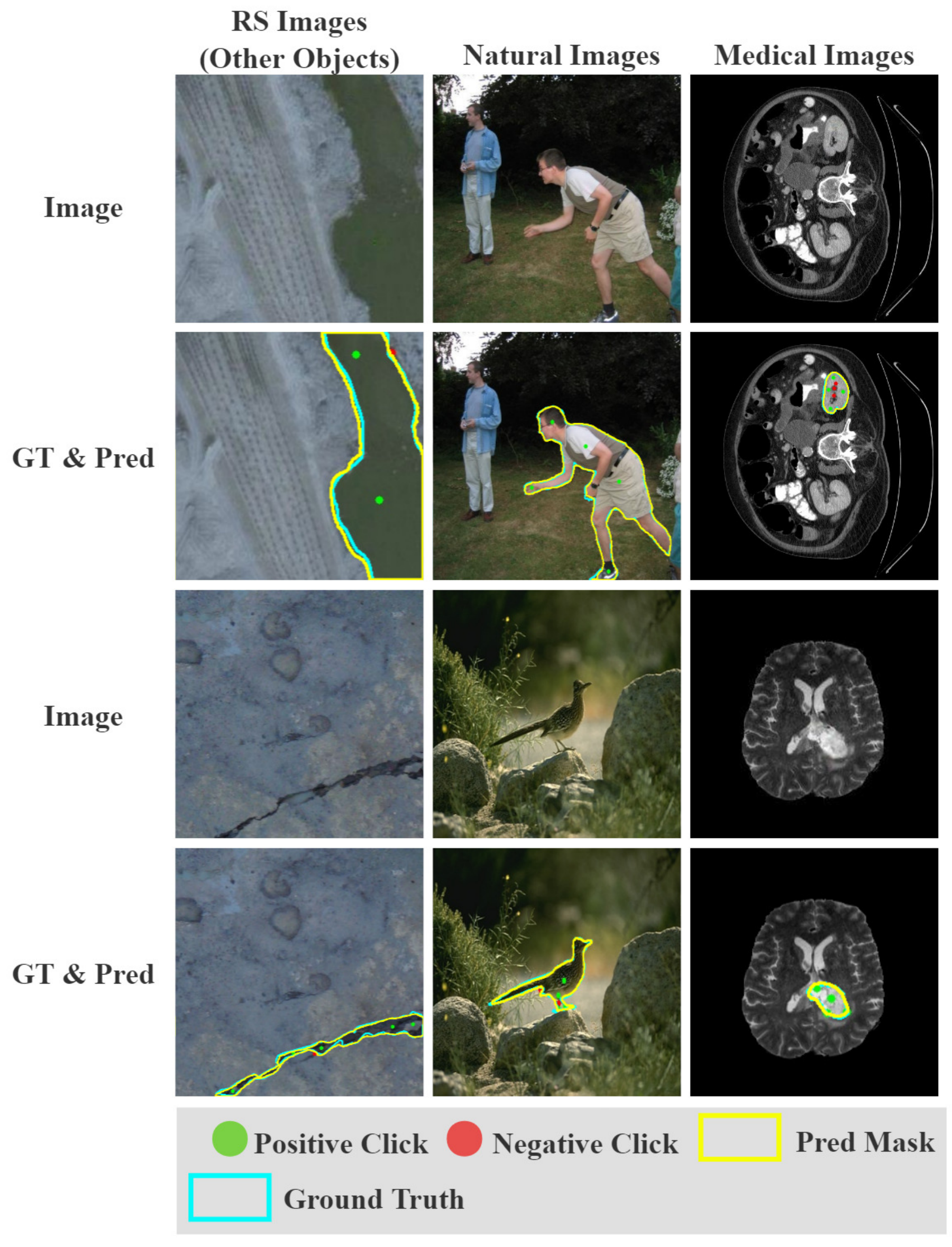}
    \caption{\textbf{Segmentation results} of the proposed model across different domains. The first column shows remote sensing images containing other ground object targets, the second column presents natural images, and the third column displays medical images, where two images are provided for each domain.}
    \label{Test_Data}
\end{figure}

\subsection{Generalizability Analysis}
This subsection explores the generalization capability of the proposed model by analyzing Fig.\ref{Test_Data}. First, our model exhibits prominent cross-domain adaptability: there are significant differences in imaging conditions among remote sensing images (e.g., rivers, surface cracks), natural images (e.g., humans, birds) and medical images (e.g., abdominal cavity, brain). Remote sensing images are characterized by large scale variations and complex backgrounds, natural images focus on the textures and contours of daily objects, and medical images demand high precision in boundary segmentation. Nevertheless, our model can generate prediction results highly consistent with the ground truth masks with no more than 8 positive or negative clicks across these three types of images. This is attributed to two key factors: first, the frozen Vision Foundation Model preserves universal visual knowledge across domains; second, the adapter module accurately captures fine-grained spatial and semantic features in different domains, thus effectively mitigating domain discrepancies.

In addition, the model demonstrates strong adaptability to target morphologies. Whether for objects with relatively regular contours in natural images (e.g., human limbs, bird trunks), irregular ground objects in remote sensing images (e.g., winding rivers, slender surface cracks) or amorphous tissues in medical images (e.g., intra-abdominal organs, complex brain structures), our model can accurately capture target boundaries relying on the adapter's ability to extract domain-specific spatial-spectral features and the probability map modulation algorithm's role in optimizing dynamic boundaries, thus avoiding overall morphological deviations or boundary blurring issues.

In summary, the proposed model exhibits excellent generalization capability in cross-domain scenarios and for diverse target morphologies, laying a solid foundation for high-precision interactive segmentation in multi-scenario applications.

\section{Conclusion}
In this work, we propose VFM-ISRefiner, a novel interactive segmentation framework specifically designed for remote sensing images. By incorporating an adapter module alongside frozen Vision Foundation Model, our method effectively balances general feature representation with domain-specific adaptation, significantly improving boundary precision. Furthermore, the integration of a convolution-transformer hybrid attention mechanism enhances the model's capability to handle large-scale variations and cluttered backgrounds, while the improved probability map modulation algorithm ensures stable propagation of historical interactions for refined segmentation. Extensive experiments on multiple RS datasets, including iSAID, SandBar, NWPU, LoveDA Urban and WHUBuilding, demonstrate that VFM-ISRefiner consistently outperforms state-of-the-art approaches in terms of segmentation accuracy, interaction efficiency, and robustness. In particular, our method achieves superior performance under high-precision settings (NoC85 and NoC90), validating its effectiveness in practical annotation scenarios.

In future work, we plan to investigate more efficient fine-tuning strategies to further reduce computational complexity, making our framework more suitable for large-scale deployment in operational remote sensing scenarios. We also aim to extend VFM-ISRefiner to more challenging settings, such as multimodal remote sensing data (e.g., hyperspectral images), large-scale annotation platforms and real-time human-computer interaction systems.

\bibliographystyle{IEEEtran}
\bibliography{reference}

\begin{thebibliography}{10}
\providecommand{\url}[1]{#1}
\csname url@samestyle\endcsname
\providecommand{\newblock}{\relax}
\providecommand{\bibinfo}[2]{#2}
\providecommand{\BIBentrySTDinterwordspacing}{\spaceskip=0pt\relax}
\providecommand{\BIBentryALTinterwordstretchfactor}{4}
\providecommand{\BIBentryALTinterwordspacing}{\spaceskip=\fontdimen2\font plus
\BIBentryALTinterwordstretchfactor\fontdimen3\font minus
  \fontdimen4\font\relax}
\providecommand{\BIBforeignlanguage}[2]{{%
\expandafter\ifx\csname l@#1\endcsname\relax
\typeout{** WARNING: IEEEtran.bst: No hyphenation pattern has been}%
\typeout{** loaded for the language `#1'. Using the pattern for}%
\typeout{** the default language instead.}%
\else
\language=\csname l@#1\endcsname
\fi
#2}}
\providecommand{\BIBdecl}{\relax}
\BIBdecl

\bibitem{xu2017deep}
N.~Xu, B.~Price, S.~Cohen, J.~Yang, and T.~Huang, ``Deep grabcut for object
  selection,'' \emph{arXiv preprint arXiv:1707.00243}, 2017.

\bibitem{sofiiuk2022reviving}
K.~Sofiiuk, I.~A. Petrov, and A.~Konushin, ``Reviving iterative training with
  mask guidance for interactive segmentation,'' in \emph{2022 IEEE
  international conference on image processing (ICIP)}.\hskip 1em plus 0.5em
  minus 0.4em\relax IEEE, 2022, pp. 3141--3145.

\bibitem{jang2019interactive}
W.-D. Jang and C.-S. Kim, ``Interactive image segmentation via backpropagating
  refinement scheme,'' in \emph{Proceedings of the IEEE/CVF conference on
  computer vision and pattern recognition}, 2019, pp. 5297--5306.

\bibitem{wu2014milcut}
J.~Wu, Y.~Zhao, J.-Y. Zhu, S.~Luo, and Z.~Tu, ``Milcut: A sweeping line
  multiple instance learning paradigm for interactive image segmentation,'' in
  \emph{Proceedings of the IEEE conference on computer vision and pattern
  recognition}, 2014, pp. 256--263.

\bibitem{liu2022survey}
P.~Liu, L.~Wang, R.~Ranjan, G.~He, and L.~Zhao, ``A survey on active deep
  learning: From model driven to data driven,'' \emph{ACM Computing Surveys
  (CSUR)}, vol.~54, no. 10s, pp. 1--34, 2022.

\bibitem{he2016deep}
K.~He, X.~Zhang, S.~Ren, and J.~Sun, ``Deep residual learning for image
  recognition,'' in \emph{Proceedings of the IEEE conference on computer vision
  and pattern recognition}, 2016, pp. 770--778.

\bibitem{liu2022convnet}
Z.~Liu, H.~Mao, C.-Y. Wu, C.~Feichtenhofer, T.~Darrell, and S.~Xie, ``A convnet
  for the 2020s,'' in \emph{Proceedings of the IEEE/CVF conference on computer
  vision and pattern recognition}, 2022, pp. 11\,976--11\,986.

\bibitem{dosovitskiy2020image}
A.~Dosovitskiy, L.~Beyer, A.~Kolesnikov, D.~Weissenborn, X.~Zhai,
  T.~Unterthiner, M.~Dehghani, M.~Minderer, G.~Heigold, S.~Gelly \emph{et~al.},
  ``An image is worth 16x16 words: Transformers for image recognition at
  scale,'' \emph{arXiv preprint arXiv:2010.11929}, 2020.

\bibitem{liu2021swin}
Z.~Liu, Y.~Lin, Y.~Cao, H.~Hu, Y.~Wei, Z.~Zhang, S.~Lin, and B.~Guo, ``Swin
  transformer: Hierarchical vision transformer using shifted windows,'' in
  \emph{Proceedings of the IEEE/CVF international conference on computer
  vision}, 2021, pp. 10\,012--10\,022.

\bibitem{song2025refining}
B.~Song, D.~Wang, P.~Liu, D.~Wang, G.~Xie, and J.~Liu, ``Refining remote
  sensing image segmentation results based on vision foundation models,''
  \emph{IEEE Geoscience and Remote Sensing Letters}, 2025.

\bibitem{liu2023simpleclick}
Q.~Liu, Z.~Xu, G.~Bertasius, and M.~Niethammer, ``Simpleclick: Interactive
  image segmentation with simple vision transformers,'' in \emph{Proceedings of
  the IEEE/CVF International Conference on Computer Vision}, 2023, pp.
  22\,290--22\,300.

\bibitem{kirillov2023segment}
A.~Kirillov, E.~Mintun, N.~Ravi, H.~Mao, C.~Rolland, L.~Gustafson, T.~Xiao,
  S.~Whitehead, A.~C. Berg, W.-Y. Lo \emph{et~al.}, ``Segment anything,'' in
  \emph{Proceedings of the IEEE/CVF international conference on computer
  vision}, 2023, pp. 4015--4026.

\bibitem{OSCO2023103540}
\BIBentryALTinterwordspacing
L.~P. Osco, Q.~Wu, E.~L. {de Lemos}, W.~N. Gonçalves, A.~P.~M. Ramos, J.~Li,
  and J.~Marcato, ``The segment anything model (sam) for remote sensing
  applications: From zero to one shot,'' \emph{International Journal of Applied
  Earth Observation and Geoinformation}, vol. 124, p. 103540, 2023. [Online].
  Available:
  \url{https://www.sciencedirect.com/science/article/pii/S1569843223003643}
\BIBentrySTDinterwordspacing

\bibitem{han2023survey}
W.~Han, X.~Zhang, Y.~Wang, L.~Wang, X.~Huang, J.~Li, S.~Wang, W.~Chen, X.~Li,
  R.~Feng \emph{et~al.}, ``A survey of machine learning and deep learning in
  remote sensing of geological environment: Challenges, advances, and
  opportunities,'' \emph{ISPRS Journal of Photogrammetry and Remote Sensing},
  vol. 202, pp. 87--113, 2023.

\bibitem{zhang2023new}
Y.~Zhang, P.~Liu, L.~Chen, M.~Xu, X.~Guo, and L.~Zhao, ``A new multi-source
  remote sensing image sample dataset with high resolution for flood area
  extraction: Gf-floodnet,'' \emph{International Journal of Digital Earth},
  vol.~16, no.~1, pp. 2522--2554, 2023.

\bibitem{wang2023interactive}
Z.~Wang, S.~Sun, X.~Que, and X.~Ma, ``Interactive segmentation in aerial
  images: a new benchmark and an open access web-based tool,'' \emph{arXiv
  preprint arXiv:2308.13174}, 2023.

\bibitem{yin2024parameter}
D.~Yin, X.~Han, B.~Li, H.~Feng, and J.~Bai, ``Parameter-efficient is not
  sufficient: Exploring parameter, memory, and time efficient adapter tuning
  for dense predictions,'' in \emph{Proceedings of the 32nd ACM International
  Conference on Multimedia}, 2024, pp. 1398--1406.

\bibitem{lester2021power}
B.~Lester, R.~Al-Rfou, and N.~Constant, ``The power of scale for
  parameter-efficient prompt tuning,'' \emph{arXiv preprint arXiv:2104.08691},
  2021.

\bibitem{awais2025foundation}
M.~Awais, M.~Naseer, S.~Khan, R.~M. Anwer, H.~Cholakkal, M.~Shah, M.-H. Yang,
  and F.~S. Khan, ``Foundation models defining a new era in vision: a survey
  and outlook,'' \emph{IEEE Transactions on Pattern Analysis and Machine
  Intelligence}, 2025.

\bibitem{rs17030369}
\BIBentryALTinterwordspacing
C.~Wei, X.~Wu, and B.~Wang, ``Ass-cd: Adapting segment anything model and
  swin-transformer for change detection in remote sensing images,''
  \emph{Remote Sensing}, vol.~17, no.~3, 2025. [Online]. Available:
  \url{https://www.mdpi.com/2072-4292/17/3/369}
\BIBentrySTDinterwordspacing

\bibitem{lin2024adaptiveclick}
J.~Lin, J.~Chen, K.~Yang, A.~Roitberg, S.~Li, Z.~Li, and S.~Li,
  ``Adaptiveclick: Click-aware transformer with adaptive focal loss for
  interactive image segmentation,'' \emph{IEEE Transactions on Neural Networks
  and Learning Systems}, vol.~36, no.~3, pp. 5759--5773, 2024.

\bibitem{lee2024mfp}
C.~Lee, S.-H. Lee, and C.-S. Kim, ``Mfp: Making full use of probability maps
  for interactive image segmentation,'' in \emph{2024 IEEE/CVF Conference on
  Computer Vision and Pattern Recognition (CVPR)}.\hskip 1em plus 0.5em minus
  0.4em\relax IEEE, 2024, pp. 4051--4059.

\bibitem{chen2024rsprompter}
K.~Chen, C.~Liu, H.~Chen, H.~Zhang, W.~Li, Z.~Zou, and Z.~Shi, ``Rsprompter:
  Learning to prompt for remote sensing instance segmentation based on visual
  foundation model,'' \emph{IEEE Transactions on Geoscience and Remote
  Sensing}, vol.~62, pp. 1--17, 2024.

\bibitem{ke2023segment}
L.~Ke, M.~Ye, M.~Danelljan, Y.-W. Tai, C.-K. Tang, F.~Yu \emph{et~al.},
  ``Segment anything in high quality,'' \emph{Advances in Neural Information
  Processing Systems}, vol.~36, pp. 29\,914--29\,934, 2023.

\bibitem{rother2004grabcut}
C.~Rother, V.~Kolmogorov, and A.~Blake, ``"grabcut" interactive foreground
  extraction using iterated graph cuts,'' \emph{ACM transactions on graphics
  (TOG)}, vol.~23, no.~3, pp. 309--314, 2004.

\bibitem{grady2006random}
L.~Grady, ``Random walks for image segmentation,'' \emph{IEEE transactions on
  pattern analysis and machine intelligence}, vol.~28, no.~11, pp. 1768--1783,
  2006.

\bibitem{boykov2001interactive}
Y.~Y. Boykov and M.-P. Jolly, ``Interactive graph cuts for optimal boundary \&
  region segmentation of objects in nd images,'' in \emph{Proceedings eighth
  IEEE international conference on computer vision. ICCV 2001}, vol.~1.\hskip
  1em plus 0.5em minus 0.4em\relax IEEE, 2001, pp. 105--112.

\bibitem{xu2016deep}
N.~Xu, B.~Price, S.~Cohen, J.~Yang, and T.~S. Huang, ``Deep interactive object
  selection,'' in \emph{Proceedings of the IEEE conference on computer vision
  and pattern recognition}, 2016, pp. 373--381.

\bibitem{sofiiuk2020f}
K.~Sofiiuk, I.~Petrov, O.~Barinova, and A.~Konushin, ``f-brs: Rethinking
  backpropagating refinement for interactive segmentation,'' in
  \emph{Proceedings of the IEEE/CVF conference on computer vision and pattern
  recognition}, 2020, pp. 8623--8632.

\bibitem{chen2022focalclick}
X.~Chen, Z.~Zhao, Y.~Zhang, M.~Duan, D.~Qi, and H.~Zhao, ``Focalclick: Towards
  practical interactive image segmentation,'' in \emph{Proceedings of the
  IEEE/CVF conference on computer vision and pattern recognition}, 2022, pp.
  1300--1309.

\bibitem{lin2020interactive}
Z.~Lin, Z.~Zhang, L.-Z. Chen, M.-M. Cheng, and S.-P. Lu, ``Interactive image
  segmentation with first click attention. in 2020 ieee,'' in \emph{CVF
  Conference on Computer Vision and Pattern Recognition (CVPR)}, vol.~10, 2020.

\bibitem{du2023efficient}
F.~Du, J.~Yuan, Z.~Wang, and F.~Wang, ``Efficient mask correction for
  click-based interactive image segmentation,'' in \emph{Proceedings of the
  IEEE/CVF Conference on Computer Vision and Pattern Recognition}, 2023, pp.
  22\,773--22\,782.

\bibitem{lin2022focuscut}
Z.~Lin, Z.-P. Duan, Z.~Zhang, C.-L. Guo, and M.-M. Cheng, ``Focuscut: Diving
  into a focus view in interactive segmentation,'' in \emph{Proceedings of the
  IEEE/CVF Conference on Computer Vision and Pattern Recognition}, 2022, pp.
  2637--2646.

\bibitem{sun2024cfr}
S.~Sun, M.~Xian, F.~Xu, L.~Capriotti, and T.~Yao, ``Cfr-icl: Cascade-forward
  refinement with iterative click loss for interactive image segmentation,'' in
  \emph{Proceedings of the AAAI conference on artificial intelligence},
  vol.~38, no.~5, 2024, pp. 5017--5024.

\bibitem{yan2023piclick}
C.~Yan, H.~Wang, J.~Liu, X.~Jiang, Y.~Hu, X.~Tang, G.~Kang, and E.~Gavves,
  ``Piclick: Picking the desired mask in clickbased interactive segmentation,''
  \emph{arXiv preprint arXiv:2304.11609}, vol.~3, 2023.

\bibitem{xu2025mst}
L.~Xu, Y.~Chen, S.~Li, J.~Luo, and Y.~Chen, ``Mst: Adaptive multi-scale tokens
  guided interactive segmentation,'' \emph{IEEE Transactions on Circuits and
  Systems for Video Technology}, 2025.

\bibitem{zhang2025ntclick}
C.~Zhang, T.~Liu, X.~Qu, L.~Liu, Y.~Zhao, and Y.~Wei, ``Ntclick: Achieving
  precise interactive segmentation with noise-tolerant clicks,'' in
  \emph{Proceedings of the Computer Vision and Pattern Recognition Conference},
  2025, pp. 8921--8930.

\bibitem{long2015fully}
J.~Long, E.~Shelhamer, and T.~Darrell, ``Fully convolutional networks for
  semantic segmentation,'' in \emph{Proceedings of the IEEE conference on
  computer vision and pattern recognition}, 2015, pp. 3431--3440.

\bibitem{simonyan2014very}
K.~Simonyan and A.~Zisserman, ``Very deep convolutional networks for
  large-scale image recognition,'' \emph{arXiv preprint arXiv:1409.1556}, 2014.

\bibitem{huang2017densely}
G.~Huang, Z.~Liu, L.~Van Der~Maaten, and K.~Q. Weinberger, ``Densely connected
  convolutional networks,'' in \emph{Proceedings of the IEEE conference on
  computer vision and pattern recognition}, 2017, pp. 4700--4708.

\bibitem{sun2019deep}
K.~Sun, B.~Xiao, D.~Liu, and J.~Wang, ``Deep high-resolution representation
  learning for human pose estimation,'' in \emph{Proceedings of the IEEE/CVF
  conference on computer vision and pattern recognition}, 2019, pp. 5693--5703.

\bibitem{vaswani2017attention}
A.~Vaswani, N.~Shazeer, N.~Parmar, J.~Uszkoreit, L.~Jones, A.~N. Gomez,
  {\L}.~Kaiser, and I.~Polosukhin, ``Attention is all you need,''
  \emph{Advances in neural information processing systems}, vol.~30, 2017.

\bibitem{he2022masked}
K.~He, X.~Chen, S.~Xie, Y.~Li, P.~Doll{\'a}r, and R.~Girshick, ``Masked
  autoencoders are scalable vision learners,'' in \emph{Proceedings of the
  IEEE/CVF conference on computer vision and pattern recognition}, 2022, pp.
  16\,000--16\,009.

\bibitem{oquab2023dinov2}
M.~Oquab, T.~Darcet, T.~Moutakanni, H.~Vo, M.~Szafraniec, V.~Khalidov,
  P.~Fernandez, D.~Haziza, F.~Massa, A.~El-Nouby \emph{et~al.}, ``Dinov2:
  Learning robust visual features without supervision,'' \emph{arXiv preprint
  arXiv:2304.07193}, 2023.

\bibitem{liang2025vision}
P.~Liang, B.~Pu, H.~Huang, Y.~Li, H.~Wang, W.~Ma, and Q.~Chang, ``Vision
  foundation models in medical image analysis: Advances and challenges,''
  \emph{arXiv preprint arXiv:2502.14584}, 2025.

\bibitem{lu2025vision}
S.~Lu, J.~Guo, J.~R. Zimmer-Dauphinee, J.~M. Nieusma, X.~Wang, S.~A. Wernke,
  Y.~Huo \emph{et~al.}, ``Vision foundation models in remote sensing: A
  survey,'' \emph{IEEE Geoscience and Remote Sensing Magazine}, 2025.

\bibitem{zhang2025high}
Z.~Zhang, X.~Hu, Y.~Yang, B.~Yang, K.~Deng, H.~Dai, and M.~Zhang,
  ``High-quality one-shot interactive segmentation for remote sensing images
  via hybrid adapter-enhanced foundation models,'' \emph{International Journal
  of Applied Earth Observation and Geoinformation}, vol. 139, p. 104466, 2025.

\bibitem{hu2022lora}
E.~J. Hu, Y.~Shen, P.~Wallis, Z.~Allen-Zhu, Y.~Li, S.~Wang, L.~Wang, W.~Chen
  \emph{et~al.}, ``Lora: Low-rank adaptation of large language models.''
  \emph{ICLR}, vol.~1, no.~2, p.~3, 2022.

\bibitem{chen2022adaptformer}
S.~Chen, C.~Ge, Z.~Tong, J.~Wang, Y.~Song, J.~Wang, and P.~Luo, ``Adaptformer:
  Adapting vision transformers for scalable visual recognition,''
  \emph{Advances in Neural Information Processing Systems}, vol.~35, pp.
  16\,664--16\,678, 2022.

\bibitem{yao2025towards}
Y.~Yao, Q.~Yang, M.~Cui, and L.~Bo, ``Towards fine-grained interactive
  segmentation in images and videos,'' \emph{arXiv preprint arXiv:2502.09660},
  2025.

\bibitem{li2024clickadapter}
S.~Li, Y.~Chen, L.~Xu, J.~Luo, R.~Huang, F.~Wu, and Y.~Miao, ``Clickadapter:
  Integrating details into interactive segmentation model with adapter,''
  \emph{IEEE Transactions on Circuits and Systems for Video Technology}, 2024.

\bibitem{zhou2023interactive}
M.~Zhou, H.~Wang, Q.~Zhao, Y.~Li, Y.~Huang, D.~Meng, and Y.~Zheng,
  ``Interactive segmentation as gaussion process classification,'' in
  \emph{Proceedings of the IEEE/CVF conference on computer vision and pattern
  recognition}, 2023, pp. 19\,488--19\,497.

\bibitem{cheng2022masked}
B.~Cheng, I.~Misra, A.~G. Schwing, A.~Kirillov, and R.~Girdhar,
  ``Masked-attention mask transformer for universal image segmentation,'' in
  \emph{Proceedings of the IEEE/CVF conference on computer vision and pattern
  recognition}, 2022, pp. 1290--1299.

\bibitem{waqas2019isaid}
S.~Waqas~Zamir, A.~Arora, A.~Gupta, S.~Khan, G.~Sun, F.~Shahbaz~Khan, F.~Zhu,
  L.~Shao, G.-S. Xia, and X.~Bai, ``isaid: A large-scale dataset for instance
  segmentation in aerial images,'' in \emph{Proceedings of the IEEE/CVF
  conference on computer vision and pattern recognition workshops}, 2019, pp.
  28--37.

\bibitem{rottensteiner2014isprs}
F.~Rottensteiner, G.~Sohn, M.~Gerke, and J.~D. Wegner, ``Isprs semantic
  labeling contest,'' \emph{ISPRS: Leopoldsh{\"o}he, Germany}, vol.~1, no.~4,
  p.~4, 2014.

\bibitem{wang2021loveda}
J.~Wang, Z.~Zheng, A.~Ma, X.~Lu, and Y.~Zhong, ``Loveda: A remote sensing
  land-cover dataset for domain adaptive semantic segmentation,'' \emph{arXiv
  preprint arXiv:2110.08733}, 2021.

\bibitem{cheng2014multi}
G.~Cheng, J.~Han, P.~Zhou, and L.~Guo, ``Multi-class geospatial object
  detection and geographic image classification based on collection of part
  detectors,'' \emph{ISPRS Journal of Photogrammetry and Remote Sensing},
  vol.~98, pp. 119--132, 2014.

\bibitem{ji2018fully}
S.~Ji, S.~Wei, and M.~Lu, ``Fully convolutional networks for multisource
  building extraction from an open aerial and satellite imagery data set,''
  \emph{IEEE Transactions on geoscience and remote sensing}, vol.~57, no.~1,
  pp. 574--586, 2018.

\bibitem{chen2021conditional}
X.~Chen, Z.~Zhao, F.~Yu, Y.~Zhang, and M.~Duan, ``Conditional diffusion for
  interactive segmentation,'' in \emph{Proceedings of the IEEE/CVF
  International Conference on Computer Vision}, 2021, pp. 7345--7354.

\end{thebibliography}

\end{document}